\documentclass[12pt,reqno]{amsart}
\usepackage{soul}
\usepackage{multicol,amsmath,graphics, mathtools,stmaryrd, cancel,soul,color,bbm,amsfonts,dsfont,tikz,mathrsfs,amssymb,bm}
\usepackage[left=1in, right=1in, top=1.1in,bottom=1.1in]{geometry}
\usetikzlibrary{matrix,patterns,positioning}
\numberwithin{equation}{section}

\usepackage{graphicx} 
\usepackage{subcaption} 
\usepackage{graphicx}
\usepackage{array}

\newcommand{\E}{\mathbb{E}}

\newcommand{\R}{\mathbb{R}}

\newcommand{\vertiii}[1]{{\left\vert\kern-0.25ex\left\vert\kern-0.25ex\left\vert #1 
    \right\vert\kern-0.25ex\right\vert\kern-0.25ex\right\vert}}

\def\R{\mathbb{R}}

\def\dh2l{\mathbf{d}_{\mathbb{H}_{2\ell}}}
\def\d2{\mathbf{d}_2}

\newtheorem{thm}{Theorem}[section]

\newtheorem{definition}[thm]{Definition}

\newtheorem{theorem}{Theorem}[section]

\theoremstyle{remark}
\newtheorem{remark}[theorem]{Remark}
\theoremstyle{definition}

\newcounter{dummy} \numberwithin{dummy}{section}

\newtheorem{Theorem}[dummy]{Theorem}

\def\1{{\rm l}\hskip -0.21truecm 1}
\begin{document}

\title[Adjusted Wasserstein distances and MDS]{Adjusted Wasserstein distances for bridging empirical and true distributions with applications to MDS}
\author{Flor Martinez-Sermeno, Arturo Jaramillo, Johan Van Horebeek}

\address{Flor Martinez-Sermeno: Department of Probability and statistics, Centro de Investigaci\'on en matem\'aticas (CIMAT)}
\email{flower10@cimat.mx}

\address{Arturo Jaramillo: Department of Probability and statistics, Centro de Investigaci\'on en matem\'aticas (CIMAT)}
\email{jagil@cimat.mx}

\address{Johan van Horebeek: Department of Computer Science, Centro de Investigaci\'on en matem\'aticas (CIMAT)}
\email{horebeek@cimat.mx}

\date{\today}

 
\begin{abstract}
This paper examines how metric adjustments to Multidimensional Scaling (MDS) can enhance its effectiveness as a visual tool for pattern recognition. The distance under consideration, referred to as Max-D-SW, is an adjustment of the Max-Sliced Wasserstein distance. In contrast to the original formulation, which optimizes over single unit directions, Max-D-SW aggregates contributions over orthonormal bases. This modification provides a clear numerical advantage in MDS outcomes, particularly when applied to heavy-tailed distributions. We also establish sample-complexity bounds showing that Max-D-SW remains statistically tractable, with rates comparable to those of its max-sliced counterpart. Moreover, we show that a better sample complexity for a metric does not necessarily translate into better performance when the metric is used as an input for MDS.
\end{abstract}

\maketitle
\section{Introduction}\label{Sec:intro}
This paper explores how a simple adjustment of the sliced Wasserstein distance can enhance the quality of data-driven visualization procedures, with a particular focus on multidimensional scaling. Our work is connected to a broader statistical theme: understanding how methods that take probability distributions as inputs behave when implemented using empirical samples. Within this broader context, we focus on how adjusted sliced-Wasserstein metrics influence multidimensional scaling, paying particular attention to the issues that arise from the mismatch between the true distribution and its empirical approximation. We begin investigating this question in a controlled manner by studying two concrete tasks: (i) comparing empirical distributions to a benchmark reference, and (ii) performing multidimensional scaling to obtain low-dimensional visual representations of data. The distance adjustments yield numerical improvements that are particularly pronounced for heavy-tailed distributions and moderately high-dimensional data (precisely the regimes where the classical max-sliced Wasserstein distance tends to degrade). We complement these controlled experiments with a simple application to image data, which illustrates how the modified metric behaves in practice. On the theoretical side, we obtain sample-complexity bounds showing that the modified distance remains statistically tractable, with rates comparable to those available for the classical formulation in the regimes considered.\\

\noindent To provide a broad context for the problem at hand, we briefly revisit how Wasserstein-type distances have been adapted in statistical applications over the past few years. A recurring theme has been the search for more tractable versions of the original Wasserstein distance, especially in high dimensions. This has led to variants like the Sliced Wasserstein distance, the Max-Sliced Wasserstein distance, and the more recent Maximum $K$-Sliced Wasserstein Distance. The main object we study here is a particular case of this latter, corresponding to the full-dimensional version $K = D$, which we refer to as the Max-$D$-Sliced Wasserstein distance.\\

\noindent \textit{Wasserstein distance and pattern recognition}\\
We begin with the $p$-Wasserstein distance, denoted by $W_p$, defined as
\begin{align*}
d_{W_p}(\mu,\nu)
&:= \left( \inf_{\pi \in \Pi(\mu, \nu)} \int_{\mathbb{R}^{d} \times \mathbb{R}^{d} } |x - y|^p , \pi(dx,dy) \right)^{1/p},
\end{align*}
where $\Pi(\mu, \nu)$ denotes the set of transport plans; that is, probability measures on $\mathbb{R}^d \times \mathbb{R}^d$ with marginals $\mu$ and $\nu$. This distance provides a natural way to compare probability distributions, and has been widely used in applications where one wants to capture how distributions shift or evolve. In our setting, we use it to describe a pattern recognition scheme designed to track how a collection of empirical distributions (data clouds) evolves over time. This idea will be explored in more detail in Section~\ref{sec:multidimensionalscaling}, where we describe how multidimensional scaling (MDS) can be used to visualize such evolution. For now, we give a rough sketch to clarify how distances are used: given a sequence of probability measures $\mu_1, \dots, \mu_m$, we compute a distance matrix $M_m$ whose entries are pairwise Wasserstein distances. We then convert this distance matrix into a centered Gram matrix and use its leading eigenvectors to obtain a lower-dimensional representation, usually in two dimensions, which often provides a good enough summary to visualize trends or transitions as the index varies.\\

\noindent Beyond this particular use case, Wasserstein distances have inspired a wide range of applications across diverse fields, including computational science, probability, optimization, and statistics. They have proven to be valuable tools in both theoretical developments \cite{villani, santambrogio} and applied methodologies. However, its practical use is often limited by significant computational demands, particularly in high-dimensional settings. This limitation is further exacerbated by the well-known curse of dimensionality (see \cite{MR236977} and \cite{MR3383341}) of the Wasserstein distance, which causes the sample complexity to grow exponentially with the dimension.\\

\noindent \textit{Sliced Wasserstein distance}\\
\noindent To address the computational challenges associated with moderate-dimensional Wasserstein distances, one popular alternative is the Sliced Wasserstein distance, which offers a more efficient approximation while preserving essential geometric features. Originally introduced in \cite{rabin} and later formalized using the Radon transform in \cite{bonneel}, this approach involves projecting empirical distributions onto one-dimensional subspaces along random directions. The $p$-Sliced Wasserstein distance is then defined by averaging the univariate Wasserstein distances between these one-dimensional projections. This yields a method that is both computationally efficient and statistically tractable, as univariate distributions are easier to estimate and the Wasserstein distance between them admits a closed-form expression (see Section \ref{eq:computationWasserstein}).\\

\noindent A known drawback of the Sliced Wasserstein distance is that it often requires a large number of projection directions to produce accurate estimates. To reduce the number of necessary projections while still capturing meaningful differences between distributions, several strategies have been proposed. A particularly influential idea is the Max-Sliced Wasserstein distance (Max-SW), introduced in \cite{deshpande2019}, which selects the single direction that maximizes the univariate Wasserstein distance between projected measures. The rationale is that most directions contribute little to distinguishing between distributions, so focusing on the most informative one can lead to better performance. While this comes at the cost of solving a nontrivial optimization problem to find the optimal direction, this is a manageable technical challenge. More importantly, there is a clear upside: recent work in \cite{nietert2022} shows that both the SW and Max-SW distances achieve convergence rates of order \( n^{-1/2} \). Although the constants involved may still depend on the dimension, the rate itself does not deteriorate with dimension, which makes these distances especially appealing for high-dimensional settings.\\

\noindent In the context of multidimensional scaling, the Sliced Wasserstein and the Max-Sliced Wasserstein distances perform well when the most significant differences between the distributions are aligned with a single dominant direction, such as a global shift or a localized deformation. However, this assumption may fail when the differences are distributed across several features. In such situations, relying on a single projection direction can obscure meaningful structural variation. To overcome this limitation, we introduce the Max-D-Sliced Wasserstein distance, which captures more complex patterns of variation. We investigate how the choice of distance influences the effectiveness of multidimensional scaling, comparing the performance of the classical Wasserstein distance, the Max-Sliced Wasserstein distance, and the Max-D-Sliced Wasserstein distance.\\

\noindent \textit{Comparison with Grassmannian-Based Distances}\\
Several recent works have proposed Wasserstein-type distances built on projections indexed by the Grassmannian manifold; which refers to the collection of all \(k\)-dimensional subspaces of \(\mathbb{R}^d\). These constructions share the same conceptual motivation as the Max-K-Sliced Wasserstein distance: they aim to enhance the expressiveness of the distance by moving from directions to higher-dimensional subspaces. We refer the reader to \cite{paty2019subspace}, \cite{lin2020projection}, and \cite{lin2022projection} for representative developments in this direction.  Both Grassmannian-based and Max-\(K\)-Sliced distances share the goal of enhancing Wasserstein-based comparisons by allowing projections into higher-dimensional subspaces. In both cases, the distance is defined through an averaging or maximization procedure within a family of subspaces. The key difference lies in the domain where this optimization occurs: in the Grassmannian-based setting, the domain is the full Grassmannian manifold, a space that is geometrically rich and highly structured. This richness, while conceptually appealing, comes at a cost: optimizing or integrating over such a space adds significant computational complexity. In contrast, the Max-D-Sliced Wasserstein distance (and, more generally, Max-\(K\) variants) operates over the set of orthonormal bases. These bases act as structured ''skeletons'' along which marginal Wasserstein distances are computed. This structural simplification plays a key role in the practical feasibility of the distances we study.\\

\noindent \textit{The Max-K-Sliced Wasserstein distance}\\
\noindent Before proceeding, we briefly review related work. The Max-K-Sliced Wasserstein distance was introduced in \cite{dai2021} within the context of a greedy deep learning algorithm. The case \(K = 1\) recovers the Max-Sliced Wasserstein distance, while \(K = D\), the ambient dimension, corresponds to the setting considered in this paper, where the largest number of orthogonal components is used.  In \cite{nguyen2023}, the authors propose a practical approximation scheme based on a greedy selection of directions. At each step \(k\), a direction \(v_k\) is chosen to maximize the Wasserstein-1 distance between the projected empirical distributions, subject to orthogonality with the previously selected directions \(v_1, \dots, v_{k-1}\). This algorithm is Markovian in the sense that it uses only the current and past directions at each step, without global optimization. While this approximation may not yield a true metric due to its heuristic nature, the theoretical Max-K-Sliced Wasserstein distance itself does satisfy the properties of a well defined distance.  In scenarios where the number of samples is smaller than the ambient dimension (which is common in deep learning mini-batch regimes), \cite{nguyen2022hierarchical} proposed the Hierarchical Sliced Wasserstein distance to better handle the issue. More recently, \cite{tran2024} proposed a Rescaled Sliced Wasserstein distance, motivated by the observation that high-dimensional data often lies in a lower-dimensional subspace. They show that, in many gradient-descent learning problems, a properly rescaled version of the standard Sliced Wasserstein distance can match or even outperform more complex alternatives.\\

\noindent It is important to notice that the contribution of our work does not reside in the formulation of the Max-D-SW distance, which has already been considered in \cite{dai2021}. Instead, the contribution resides in pointing out, and understanding via numerical examples, that its effect could be quite powerful in pattern recognition problems. The rest of the paper is organized as follows. Section 2 reviews the necessary preliminaries, including the Wasserstein distance, its sliced and max-sliced variants, basic tools from empirical process theory, and the use of multidimensional scaling in pattern recognition. Section 3 presents our main contributions: we establish theoretical properties of the Max-D-Sliced Wasserstein distance, derive sample complexity bounds, and illustrate its advantages through numerical experiments, particularly in tasks involving heavy-tailed distributions and data embedded in higher-dimensional spaces. Section 4 concludes with final remarks and future research directions.

\section{Preliminaries}
\subsection{Computation of the Wasserstein Distance}\label{eq:computationWasserstein}
\noindent In general, computing the Wasserstein distance can be computationally demanding. However, in one dimension, the problem becomes significantly more tractable thanks to a closed-form expression involving quantile functions. Given a probability measure $\mu$ on $\mathbb{R}$, we denote its quantile function by $F_\mu^{-1}$, highlighting its role as the right-inverse of the cumulative distribution function. In the sequel, for a function $f:\R^{d}\rightarrow\R$, and a probability measure $\rho$ over $\R^{d}$, we will denote by $f_{\#}\rho$ the associated push-forward measure. A classical result states that pushing forward the Lebesgue measure $\lambda$ on $[0,1]$ through $F_\mu^{-1}$ yields $\mu$, that is, $(F_\mu^{-1})_{\#}\lambda = \mu$. This fact allows one to express the $p$-Wasserstein distance between two probability measures $\mu$ and $\nu$ as
\begin{equation}\label{wassuni}
    d_{W_p}(\mu, \nu) = \left( \int_0^1 \lvert F_\mu^{-1}(x) - F_\nu^{-1}(x) \rvert^p \, dx \right)^{1/p},
\end{equation}
providing a simple and efficient way to compute it in the univariate case.  The identity in \eqref{wassuni} leads to a simple expression for the Wasserstein distance between empirical measures. Suppose $\hat{\mu}$ and $\hat{\nu}$ are the empirical distributions associated with samples $X = \{x_1, \dots, x_n\}$ and $Y = \{y_1, \dots, y_n\}$, both of size $n$. Then, the $p$-Wasserstein distance between them can be written as
\begin{equation}\label{wassorden}
    d_{W_p}(X, Y) = \left( \dfrac{1}{n} \sum_{i=1}^n \lvert x_{(i)} - y_{(i)} \rvert^p \right)^{1/p},
\end{equation}
where $x_{(i)}$ and $y_{(i)}$ denote the $i$-th order statistics of $X$ and $Y$, respectively.
		
\subsection{Sliced Wasserstein distance}\label{sec:slicedandmaxsliced}
Taking advantage of the closed formula for the Wasserstein distance in one dimension,  \cite{rabin} proposed the Sliced Wasserstein distance projecting over random directions for discrete distributions and then \cite{bonneel} formalized it using the Radon transform. The idea behind the $p$-Sliced Wasserstein distance is to combine univariate Wasserstein distances between the marginals of the multivariate distributions, instead of computing the Wasserstein distance between them. The following material follows closely \cite{kolourigensliced}.\\

\noindent  For a given element $\theta$ in the $(d-1)$-dimensional sphere $\mathbb{S}^{d-1}$, we will denote by $R^{\theta}:\R^{d}\rightarrow\R$ the projection function $R^{\theta}(x):=\theta\cdot x.$  The $p$-Sliced Wasserstein distance between $\mu$ and $\nu$ is defined as
		\begin{equation}
			d_{SW_p}(\mu, \nu) 
			= \left(\int_{\mathbb{S}^{d-1}} |d_{W_p}(R_{\#}^{\theta}[\mu],R_{\#}^{\theta}[\nu])|^pd\theta\right)^{1/p}\label{swdef}.
		\end{equation}		
In practice, this allows to use a Monte Carlo scheme to approximate the intractable integral over $\mathbb{S}^{d-1}$, 
 sampling values for $\theta$ and computing the distance. The distributions are approximated by their empirical estimator and the distance with the Wasserstein distance between them. Therefore, to calculate the $p$-Sliced Wasserstein distance in practice, it is necessary to compute the projection of empirical distributions. Although the $p$-Sliced Wasserstein distance has good sample complexity properties and is not computationally expensive, it requires a lot of projection directions to be correctly estimated. This property, known as projection complexity, complicates its use in high dimensional settings, reason for which a lot of effort has been put in improving how the projection directions are chosen \cite{deshpande2019, nguyen2020, nguyen2022hierarchical, nguyen2023, nguyenenergy}.

\subsection{Max-sliced Wasserstein distance}\label{sec:maxslicedandmaxsliced}
A similar logic that drives us to propose the sliced Wasserstein distance can be implemented by replacing the integration operation by a maximum operation, which gives rise to the max-sliced Wasserstein distance, whose definition we discuss next. The Max Sliced Wasserstein distance between two probability measures $\mu$ and $\nu$  in $\R^{d}$, is defined as
	\begin{equation}
	d_{MaxSW}(\mu, \nu)
		= \max_{\theta \in \mathbb{S}^{d-1}} d_{W_p}\left(R_{\#}^{\theta}[\mu], R_{\#}^{\theta}[\nu] \right).
	\end{equation}
This distance has been successfully used in learning settings, and has the same sample complexity as the Sliced Wasserstein distance and has shown empirically to have improved sensitivity properties in comparison with the Sliced Wasserstein distance \cite{deshpande2019, nguyenenergy, nguyen2023}. This makes it a very strong candidate for being chosen as a good input for the multidimensional scaling.  Regardless of this, in Section \ref{sec:maincontributions}, we exhibit natural instances (mostly consisting of cases where multiple important differences along multiple different directions within the data are expected), where implementing a projection over a single element in $\mathbb{S}^{d-1}$ may not be enough. A natural option to address this problem would be to consider Grassmannian distances to consider different directions, but the price to pay would be to gain a strong increase in the sample complexity when the dimensionality of the data is large. Thus, in order to maintain a compelling projection complexity and at the same time retaining more than one direction, one naturally arrives to considering the Max-K-Sliced Wasserstein distance. Moreover, due to the fact that we are aiming to consider as many directions as possible, we exclusively focus in the case where $K$ is the full dimension, leading to the use of the Max-D-Sliced Wasserstein distance.

\subsection{Empirical Processes}
We now discuss the main tool to be used for purposes of bounding sample complexity: the empirical process theory. This section summarizes the key concepts we will need, following closely the notation and presentation in \cite{MR4628026}. Our main objective is to establish Theorem \ref{thmmain:empirical}, which provides concrete bounds on the estimation of the discrepancy between an empirical distribution and its underlying generating probability measure. To state it precisely, we begin with a few definitions.\\

\noindent \textit{Basic definitions}\\
\noindent Let $\mathcal{F}$ be a collection of measurable functions from $\R^{d}$ to $\R$. Consider as well a  collection $\xi_1,\dots, \xi_r$ of independent random variables with common distribution $\mu$, satisfying that $f$ is integrable with respect to $\mu$ for every $f$ belonging to $\mathcal{F}$. We will assume that these random variables, along with the ones mentioned in the sequel, are defined over a common probability space $(\Omega,\mathcal{G},\mathbb{P})$. These ingredients allow us to define the process $\mathbb{G}_r=\{\mathbb{G}_r[f]\ ;\ f\in\mathcal{F}\}$, with 
\begin{align*}
\mathbb{G}_r[f]
  &:=\sqrt{r}\left(\int_{\R^{d}}f(x)\hat{\mu}_r(dx)-\int_{\R^{d}}f(x)\mu(dx)\right),	
\end{align*}
where $\hat{\mu}_r$ denotes the empirical distribution associated to the cloud of points $\xi_{1},\dots, \xi_r$. For convenience, we will assume that the mapping  $\omega\mapsto \sup_{f\in\mathcal{F}}|\mathbb{G}_r[f](\omega)|$
is measurable. In this case, we define
\begin{align*}
\|\mathbb{G}_{r}\|_{\mathcal{F}}
  &:=\E[\sup_{f\in\mathcal{F}}|\mathbb{G}_r[f]|].
\end{align*}

\noindent \textit{Bracketing number}\\
\noindent Given two elements $f_1, f_2$ in $ \mathcal{F}$, we define the bracket $[f_1, f_2]$ as the set of functions $f$ in $ \mathcal{F}$, such that $f_1 \leq f \leq f_2$. Based on this, we define the {bracketing number} of the class $\mathcal{F}$ at level $\varepsilon > 0$, denoted by $N_{[\,]}(\varepsilon, \mathcal{F})$, as
\begin{align*}
N_{[\,]}(\varepsilon,\mathcal{F})
	&:= \min \left\{ N \in \mathbb{N} \ ;\ \text{there exist } (a_1, b_1), \dots, (a_N, b_N) \in \mathcal{D}_\varepsilon \text{ such that } \mathcal{F} \subset \bigcup_{i=1}^N [a_i, b_i] \right\},
\end{align*}
where $\mathcal{D}_\varepsilon$ denotes the set of pairs $(a, b)$ in $\mathcal{F}^2$ satisfying $\|b - a\|_{L^2(\mu)} < \varepsilon$.  Strictly speaking, this definition deviates slightly from the general setup in \cite{MR4628026}, where the norm used to define the brackets is arbitrary. For clarity and consistency, we restrict ourselves to the $L^2(\mu)$-norm throughout, with the purpose of avoiding technical complications that arise when working with more general norms. The following theorem, presented as Theorem 2.14.2 in \cite{MR4628026} is the main result in this section 

\begin{Theorem}[Theorem 2.14.2 in \cite{MR4628026}]\label{thmmain:empirical}
Suppose that \(F\) is an envelope for \(\mathcal{F}\) and that \(F\in L^2(\mu)\). There exists a universal constant \(C>0\) such that 
\begin{align*}
\|\mathbb{G}_{r}\|_{\mathcal{F}}
  &\leq C\|F\|_{L^2(\mu)}
  \int_0^{1}\sqrt{1+\log\left(N_{[\,]}\left(\varepsilon\|F\|_{L^{2}(\mu)},\mathcal{F}\right)\right)}\,d\varepsilon.
\end{align*}	
\end{Theorem}
The above result will be crucial for estimating the sample complexity of different estimates regarding the Max-D-Sliced Wasserstein.\\

\noindent \textit{Covering number}\\
\noindent As one might expect, estimating the bracketing number is essential for applying the result above. While several approaches are available, in this paper we adopt a perspective based on covering numbers, defined as follows. Let \( (T, d) \) be a metric space. For \( \varepsilon > 0 \), the {covering number} \( N(\varepsilon, T, d) \) is the minimal number of closed balls of radius \( \varepsilon \) (with respect to the metric \( d \)) needed to cover \( T \). It is a standard fact that if \( T \) is a compact \( \ell \)-dimensional smooth manifold, then there exists a universal constant \( C > 0 \) such that
\[
\log N(\varepsilon, T, d) \leq C \ell \log(1/\varepsilon),
\]
for all sufficiently small \( \varepsilon > 0 \). This reflects the fact that compact manifolds behave, at small scales, like subsets of \( \mathbb{R}^{\ell} \). The connection of this definition to the bracketing numbers is formulated by the following theorem

\begin{Theorem}[Theorem 2.7.11 in \cite{MR4628026}]\label{thm2711}
Let \( \mathcal{F} = \{ f_t : t \in T \} \) be a class of functions satisfying
\[
|f_s(x) - f_t(x)| \leq d(s, t) \cdot F(x),
\]
for $s,t\in T$, for some function \( F \). Then,
\[
N_{[\,]}\left( 2c \|F\|_{L^{2}(\mu)}, \mathcal{F} \right) \leq N(c, T, d).
\]
\end{Theorem}

\subsection{Multidimensional Scaling}\label{sec:multidimensionalscaling} 
In many applications, including tasks in classification and visualization, one wishes to encode complex objects (such as probability measures) as points in a Euclidean space so that standard geometric tools become applicable. Multidimensional scaling (MDS) offers a classical way to achieve this. The procedure begins with a matrix of pairwise dissimilarities between the objects under consideration and produces an embedding into \( \mathbb{R}^k \) in which the resulting Euclidean distances approximate the original dissimilarities as closely as possible.  For foundational references and further discussion, see \cite{MR2158691}. We summarize the basic steps of the method below.\\

\noindent
Consider a collection of \( r \) objects in a metric space, and let  
\( D = (d_{ij}) \) be the \( r \times r \) matrix of pairwise distances between them.  
Let \( D^{(2)} = (d_{ij}^2) \) denote the matrix of squared distances, and define the centering matrix  
\[
J = I_r - \frac{1}{r}\,\mathbf{1}\mathbf{1}^\top .
\]
The {centered Gram matrix} associated with \(D\) is then defined by
\[
B := -\frac{1}{2}\, J D^{(2)} J.
\]

\noindent
To obtain an explicit embedding, we perform a spectral decomposition of the form $B=U \Lambda U^\top$, where \( \Lambda = \mathrm{diag}(\lambda_1,\dots,\lambda_r) \) contains the eigenvalues in non-increasing order, and  
\( U = [u_1 \ \cdots \ u_r] \) is the matrix of corresponding orthonormal eigenvectors.  
For a prescribed embedding dimension \( k \le r \), we retain the leading positive eigenvalues. The embedded points  
\( x_1,\dots,x_r \in \mathbb{R}^k \) are then defined by
\[
x_i := \left( \sqrt{\lambda_1}\,u_{i1},\ \dots,\ \sqrt{\lambda_k}\,u_{ik} \right),
\]
where \(u_{ij}\) denotes the \(i\)-th coordinate of the \(j\)-th eigenvector.

\section{Main contributions}\label{sec:maincontributions}
\noindent This section is devoted to the study of the computational performance of the Max-D-SW distance, along with some theoretical properties, which constitute the main contributions of our paper.	 In the sequel, \(\mathbb{V}(\mathbb{R}^d)\) will denote the set of all orthonormal bases of \(\mathbb{R}^d\). For $v$ in  \(\mathbb S^{d-1}\), we write
\[
\mu^v := (R^v)_\#\mu.
\]
We start with the definition of the Max-D-SW distance, obtained as a specialization of the framework introduced in \cite{dai2021}.
\begin{definition}\label{maxdsw}
The Max-D-Sliced Wasserstein distance between $\mu$ and $\nu$ is defined as:
\begin{equation}\label{eq:defMaxdSW}
		d_{\text{MaxDSW}}(\mu, \nu) = \max_{\{v_1,...,v_d\} \in \mathbb{V}(\mathbb{R}^d)} \sum_{i=1}^{d} d_{W_p}\left(\mu^{v_i}, \nu^{v_i} \right).
\end{equation}
\end{definition}

\noindent We have avoided the dependence on $p$ for ease of notation, although the reader should keep in mind that this dependence is present. The Max-D-Sliced Wasserstein distance induces the same topology as the standard Wasserstein distance. This can be seen through a direct comparison with the usual Max-Sliced Wasserstein distance. Define
\[
d_{\mathrm{MaxSW},p}(\mu,\nu)
:=
\sup_{v\in \mathbb S^{d-1}}
d_{W_p}\big((R^v)_\#\mu,(R^v)_\#\nu\big),
\]
where \(R^v(x)=v\cdot x\). Since every direction \(v\in\mathbb S^{d-1}\) can be completed to an orthonormal basis of \(\mathbb R^d\), we have
\[
d_{\mathrm{MaxSW},p}(\mu,\nu)
\le
d_{\mathrm{MaxDSW},p}(\mu,\nu).
\]
On the other hand, for every orthonormal basis \((v_1,\dots,v_d)\),
\[
\sum_{i=1}^d
d_{W_p}\big((R^{v_i})_\#\mu,(R^{v_i})_\#\nu\big)
\le
d\, d_{\mathrm{MaxSW},p}(\mu,\nu).
\]
Taking the maximum over all orthonormal bases gives
\[
d_{\mathrm{MaxSW},p}(\mu,\nu)
\le
d_{\mathrm{MaxDSW},p}(\mu,\nu)
\le
d\, d_{\mathrm{MaxSW},p}(\mu,\nu).
\]
Hence \(d_{\mathrm{MaxDSW},p}\) and \(d_{\mathrm{MaxSW},p}\) are equivalent up to constants depending only on the dimension. Since the Max-Sliced Wasserstein distance is known to induce the same topology as the \(p\)-Wasserstein distance on \(\mathcal P_p(\mathbb R^d)\) (see \cite{nietert2022}), the same conclusion follows for \(d_{\mathrm{MaxDSW},p}\). In particular, this distance preserves the same notion of convergence while allowing the comparison to aggregate information over a full orthonormal basis rather than over a single optimal direction.\\

\noindent  Definition \ref{maxdsw} is clearly motivated by the definition of the Max-Sliced Wasserstein distance, but as mentioned in the introduction, the adjustment aims to deal with the fact that in situations where a ``movement'' over more than one projection direction is experienced in the application in hand, important information over some of the other directions might be lost. The practical effects of this are discussed in detail in the present section from a  numerical point of view, by exhibiting examples where substantial differences in performance are drastic. From the theoretical side, we show sample-complexity bounds that are comparable to the known bounds for the max-sliced distance, supporting the idea that the adjustment in the distance considered here still yields a statistically tractable object. These two sections are the main contributions of our paper.\\

\noindent Before delving further into the details of this, we would like to mention that in the authors opinion, it is extremely surprising that numerical improvements so radical are obtained by means of such a small adjustment in the distance (from Max-Sliced to Max-D-SW). We have not found yet a theoretical property that could explain this, but by looking through the lens of sample complexity, we at least show that the resulting distance remains statistically tractable.\\   

\noindent In the following, we present an analysis of the sample complexity associated with the distances we study. Specifically, we investigate how large a sample is required for the empirical measure to approximate the true distribution in expectation. Given a tolerance \( \varepsilon > 0 \), our goal is to determine a lower bound \( L \) such that for all \( n \geq L \), and for any sample \( \xi_1, \dots, \xi_n \) of independent and identically distributed random variables with common law \( \mu \) and empirical distribution \( \hat{\mu}_n \), the following inequality holds:
\begin{align*}
\E\big[d_{\mathrm{MaxDSW}}(\hat{\mu}_n,\mu)\big] \leq \varepsilon.
\end{align*}

\noindent We focus on the case where the Max-D-SW distance is computed with power \( p = 1 \). In this setting, the dual formulation of the Wasserstein distance is given by the Kantorovich-Rubinstein expression:
\begin{align}\label{eq:Kantorovichidentity}
d_{W_1}(\rho_1, \rho_2) = \sup_{f \in \text{Lip}_1(\mathbb{R})} \left| \int_{\mathbb{R}} f(x)\, \rho_1(dx) - \int_{\mathbb{R}} f(x)\, \rho_2(dx) \right|,
\end{align}
where \( \rho_1 \) and \( \rho_2 \) are probability measures on \( \mathbb{R} \) with finite first moments. Since both \( \rho_1 \) and \( \rho_2 \) are probability measures, subtracting a constant such as \( f(0) \) from the test function leaves the integral difference unchanged. Hence, we may restrict the supremum to one-Lipschitz functions vanishing at zero, without affecting the value of the distance.\\

\noindent Let \(\mathcal F^0\) denote the class of one-Lipschitz functions on \(\mathbb R\) which vanish at zero. For \(M>0\), let \(\mathcal F_M^0\) denote the class of one-Lipschitz functions on \([-M,M]\) which vanish at zero. We define \(\mathcal F\) as the class of functions of the form
\[
f_{(v,g)}(x):=g(v\cdot x),
\]
for $v$ in $\mathbb S^{d-1}$ and $ g$ in $\mathcal F^0.$ For the truncated class, let
\[
\pi_M(u):=(-M)\vee u\wedge M
\]
and define \(\mathcal F_M\) as the class of functions of the form
\[
f_{(v,g)}(x):=g(\pi_M(v\cdot x)),
\]
for $ v$ in $\mathbb S^{d-1}$ and $g$ in $\mathcal F_M^0.$ Clearly, the function \(F(x):=1+|x|\) is an envelope for the classes considered below. Under the heavy-tail assumption with \(\beta>2\), this envelope belongs to \(L^2(\mu)\), which allows us to apply the bracketing estimate below. Let $\mathbb{G}_n$ denote the empirical process associated to the sample $\xi_1,\dots, \xi_n$ that generates $\hat{\mu}_n$. By the representation \eqref{eq:Kantorovichidentity}, 
\begin{align*}
d_{\mathrm{MaxDSW}}(\mu,\nu)
  &= \max_{\{v_1,...,v_d\} \in \mathbb{V}(\mathbb{R}^d)} 
      \sum_{i=1}^d \sup_{g\in\mathcal{F}^0}
        \left|\int_{\R^{d}} g(v_i\cdot x)\,(\mu-\nu)(dx)\right| \\
  &\leq  d\,
     \sup_{v\in\mathbb{S}^{d-1}}\sup_{g\in\mathcal{F}^0}
        \left|\int_{\R^{d}} g(v\cdot x)\,(\mu-\nu)(dx)\right|.
\end{align*}
From the definition of $\mathcal{F}$, it then follows that
\[
d_{\mathrm{MaxDSW}}(\mu,\nu)
  \le d\,\sup_{h\in\mathcal{F}}
        \left|\int_{\R^{d}} h(x)\,(\mu-\nu)(dx)\right|.
\]
In particular, if $\hat{\mu}_n$ is the empirical measure of an i.i.d.\ sample
$\xi_1,\dots,\xi_n\sim\mu$, and we denote by
\[
\mathbb{G}_n(h)
  := \frac{1}{\sqrt{n}}
     \sum_{i=1}^n
       \bigl(h(\xi_i)-\mathbb{E}[h(\xi_1)]\bigr),
\]
for $h$ in $\mathcal{F},$ the associated empirical process, then
\[
\mathbb{E}\bigl[d_{\mathrm{MaxDSW}}(\hat{\mu}_n,\mu)\bigr]
  \le \frac{d}{\sqrt{n}}\,
      \|\mathbb{G}_n\|_{\mathcal{F}}.
\]
By elementary algebraic manipulations we then get
\begin{align}\label{eq:Expecationdmaxdsw}
\mathbb{E}\big[d_{\mathrm{MaxDSW}}(\hat{\mu}_n,\mu)\big]
  &\le \frac{d}{\sqrt{n}}\,\|\mathbb{G}_n\|_{\mathcal{F}_M}
     + 2d\,\mathbb{E}\big[|\xi|\,\mathbf{1}_{\{|\xi|>M\}}\big],
\end{align}
where $\xi$ denotes a random variable with distribution $\mu$. To control the second term, it suffices to bound the tail probability of  \( {\mu}\). To this end, we will distinguish two fundamentally different cases: (i) the case where $\mu$ has moderately heavy tails and when it has subexponential tails.\\

\noindent We start with the heavy-tail case and assume that the law $\mu$ satisfies
\begin{align}\label{eq:heavytailestimate}
\mu\big[\{x\in\mathbb R^d: |x|>t\}\big]
  \leq \frac{\kappa}{t^\beta},
\end{align}
for all \(t \geq t_0\), and some constants \(\kappa>0\), \(\beta>2\), and \(t_0>0\). Using the tail-integral identity for nonnegative random variables, we have
\[
\mathbb{E}\big[|\xi|\,\mathbbm{1}_{\{|\xi|>M\}}\big]
  =
  M\,\mu\big[\{x\in\mathbb R^d: |x|>M\}\big]
  +
  \int_M^\infty \mu\big[\{x\in\mathbb R^d: |x|>t\}\big]dt.
\]
Thus, by \eqref{eq:heavytailestimate},
\[
\mathbb{E}\big[|\xi|\,\mathbbm{1}_{\{|\xi|>M\}}\big]
  \leq
  \kappa M^{1-\beta}
  +
  \int_M^\infty \frac{\kappa}{t^\beta}\,dt
  =
  \frac{\kappa\beta}{\beta-1}M^{1-\beta}.
\]
Combining this with \eqref{eq:Expecationdmaxdsw}, we conclude that for $M\ge t_0$,
\begin{align}\label{ineq:boundontail}
\mathbb{E}\big[d_{\mathrm{MaxDSW}}(\hat{\mu}_n,\mu)\big]
  &\le \frac{d}{\sqrt{n}}\,\|\mathbb{G}_n\|_{\mathcal{F}_M}
     + C d\,M^{1-\beta}.
\end{align}
By Theorem~\ref{thmmain:empirical}, it holds that 
\begin{align*}
\mathbb{E}\big[d_{\mathrm{MaxDSW}}(\hat{\mu}_n,\mu)\big]
  \leq
  \frac{C d\|F\|_{L^2(\mu)}}{\sqrt{n}}
  \int_0^{1}
  \sqrt{
  1+\log\left(
  N_{[\,]}\left(\varepsilon\|F\|_{L^{2}(\mu)},\mathcal{F}_M\right)
  \right)}
  \,d\varepsilon
  + C d\,M^{1-\beta}.
\end{align*}
We now verify that the class \(\mathcal F_M\) satisfies the hypothesis of Theorem~\ref{thm2711}. More precisely, we will use a product metric \(d_T\) on the parameter space \(T\) for which
\begin{align*}
|f_s(x)-f_t(x)|
  &\leq d_T(s,t)F(x),
\end{align*}
with envelope \(F(x):=1+|x|\). Recall that \(\pi_M\) denotes the truncation map onto \([-M,M]\). For the truncated class, write
\[
f_{(v,g)}(x):=g(\pi_M(v\cdot x)),
\qquad (v,g)\in T:=\mathbb S^{d-1}\times\mathcal F_M^0,
\]
where \(\mathcal F_M^0\) denotes the class of one-Lipschitz functions on \([-M,M]\) which vanish at zero. We equip \(T\) with the product metric
\[
d_T((v,g),(w,h))
:=
|v-w|+\|g-h\|_{\infty,[-M,M]}.
\]
Then, for \(s=(v,g)\) and \(t=(w,h)\), we have
\begin{align*}
|f_s(x)-f_t(x)|
&\leq
|g(\pi_M(v\cdot x))-g(\pi_M(w\cdot x))|
+
|g(\pi_M(w\cdot x))-h(\pi_M(w\cdot x))| \\
&\leq
|v-w|\,|x|+\|g-h\|_{\infty,[-M,M]}
\leq
d_T(s,t)(1+|x|).
\end{align*}
Thus the desired estimate follows with \(F(x)=1+|x|\). Consequently, by Theorem~\ref{thm2711}, the bracketing number on the right can be estimated as 
\begin{align*}
N_{[\,]}\left(\varepsilon\|F\|_{L^{2}(\mu)},\mathcal{F}_M\right)
  \leq N(\varepsilon/2,T,d_T),
\end{align*}
where the right-hand side denotes the associated covering number of \(T\) with respect to the metric \(d_T\). Observe that if $D_1^1,\dots, D_{r_1}^1$ is a covering for $\mathbb{S}^{d-1}$ and $D_1^2,\dots, D_{r_2}^2$ is a covering for $\mathcal{F}^0_M$, then the sets $D_i^{1}\times D_j^{2}$, with $i\in[r_1]$ and $j\in[r_2]$, form a covering for $T$. In particular, 
\begin{align*}
\log N_{[\,]}\left(\varepsilon\|F\|_{L^{2}(\mu)},\mathcal{F}_M\right)
  &\leq \log N(\varepsilon/2,\mathbb{S}^{d-1})
        + \log N(\varepsilon/2,\mathcal{F}^0_M).
\end{align*}

\noindent Since \(\mathbb S^{d-1}\) is a smooth compact manifold of dimension \(d-1\), its covering number satisfies
\[
\log N(\varepsilon,\mathbb S^{d-1})
\leq C(d-1)\log(1/\varepsilon)
\]
for all sufficiently small \(\varepsilon>0\). On the other hand, the class \(\mathcal F_M^0\) of one-Lipschitz functions on \([-M,M]\) vanishing at zero satisfies the standard entropy bound for Lipschitz classes; see, for instance, the entropy estimates for H\"older classes in \cite[Section 2.7]{MR4628026}. Hence
\[
\log N(\varepsilon,\mathcal F_M^0,\|\cdot\|_{\infty,[-M,M]})
\leq C\frac{M}{\varepsilon},
\]
for a universal constant \(C>0\). Combining this with the covering estimate for \(\mathbb S^{d-1}\), we obtain the desired bound for the product parameter space \(T\). Taking \(M=n^\vartheta\), for \(\vartheta\geq 0\), we obtain
\begin{align*}
\E[d_{\mathrm{MaxDSW}}(\hat{\mu}_n,\mu)]
  &\leq \kappa d\left(n^{-\vartheta(\beta-1)}
  +\frac{1}{\sqrt{n}}\int_0^{1}
  \sqrt{1+\left(1+\kappa (d-1)\log(1/\varepsilon)+n^{\vartheta}/\varepsilon\right)}
  \,d\varepsilon\right),
\end{align*}
for some constant \(\kappa>0\). Using Young's inequality to split the terms inside the square root, and using that
\[
\int_0^1 \varepsilon^{-1/2}\,d\varepsilon<\infty,
\]
we obtain the bound
\[
\E[d_{\mathrm{MaxDSW}}(\hat{\mu}_n,\mu)]
  \leq
  C d^{3/2}
  \left(
  n^{-\vartheta(\beta-1)}
  +
  n^{(\vartheta-1)/2}
  \right).
\]
Balancing the two powers gives
\[
\vartheta=\frac{1}{2\beta-1}.
\]
Therefore,
\begin{align*}
\E[d_{\mathrm{MaxDSW}}(\hat{\mu}_n,\mu)]
  &\leq
  C d^{3/2}
  n^{-(\beta-1)/(2\beta-1)}.
\end{align*}
In the case where $\mu$ is subexponential, we obtain the following inequality instead of \eqref{ineq:boundontail}
\begin{align}\label{ineq:boundontail2}
\E[d_{\mathrm{MaxDSW}}(\hat{\mu}_n,\mu)]
  &\leq
  \frac{d}{\sqrt{n}}\|\mathbb{G}_n\|_{\mathcal{F}_M}
  + C d e^{-\delta M},
\end{align}
for some $\delta,C>0$. Following an argument analogous to the heavy tailed case, but replacing the choice of $M$ by a multiple constant of $\log(n)$, we obtain that 
\begin{align*}
\E[d_{\text{MaxDSW}}(\hat{\mu}_n,\mu)]
  &\leq Cd^{3/2}\log(n)n^{-1/2}.
\end{align*}
The following result summarizes the previous discussion and constitutes our main theoretical contribution.

\begin{Theorem}\label{eq:theorem_main_paper}
Denote by \(\psi_{\mu}:\mathbb R_{+}\rightarrow\mathbb R\) the tail function
\[
\psi_{\mu}(z):=\mu\big[\{x\in\mathbb R^d: |x|>z\}\big].
\]

\begin{enumerate}
\item[-] If there exists \(C>0\) such that
\[
\psi_{\mu}(z)\leq C(1+z)^{-\beta},
\qquad z\geq 0,
\]
for some \(\beta>2\), then there exists a constant \(\kappa>0\) such that
\begin{align*}
\E[d_{\mathrm{MaxDSW}}(\hat{\mu}_n,\mu)]
  &\leq
  \kappa d^{3/2} n^{-(\beta-1)/(2\beta-1)}.
\end{align*}

\item[-] If there exist constants \(C,\delta>0\) such that
\[
\psi_{\mu}(z)\leq C e^{-\delta z},
\]
for $z\geq 0$, then there exists \(\kappa>0\) such that 
\begin{align}\label{eq:Expmainthm}
\E[d_{\text{MaxDSW}}(\hat{\mu}_n,\mu)]
  &\leq \kappa d^{3/2}\log(n)n^{-1/2}.
\end{align}	
\end{enumerate}
In particular, the sample complexity of $d_{\text{MaxDSW}}(\hat{\mu}_n,\mu)$ is upper bounded by $\Upsilon(\varepsilon,d,\mu)$, where
 \begin{align}\label{eq:upsilonedmu}
\Upsilon(\varepsilon,d,\mu)
  &:=\left\{\begin{array}{lll}
(\kappa d^{3/2}/\varepsilon)^{(2\beta-1)/(\beta-1)}
&\text{ if }&
\psi_{\mu}(z)\leq C(1+z)^{-\beta}\text{ for some }\beta>2,\\[0.2cm]
(\kappa d^{3/2}\log(d/\varepsilon)/\varepsilon)^{2}
&\text{ if }&
\psi_{\mu}(z)\leq C e^{-\delta z}\text{ for some }C,\delta>0.
\end{array}\right.
 \end{align}
\end{Theorem}

\begin{remark}
The above bound is of the same order as the ones already considered in the literature, and, up to a universal multiple constant, a good performance for estimations for $d_{\text{MaxDSW}}(\hat{\mu}_n,\mu)$ requires the same sample size as the max-SW distance.
\end{remark}

\noindent Theorem \ref{eq:theorem_main_paper} directly addresses the problem of quantifying how well 
\( d_{\text{MaxDSW}}(\hat{\mu}_n,\hat{\nu}_n) \) approximates 
\( d_{\text{MaxDSW}}(\mu,\nu) \). To connect the theorem with this approximation question, observe that the reverse triangle inequality gives
\[
\big|d_{\text{MaxDSW}}(\hat{\mu}_n,\hat{\nu}_n)
      - d_{\text{MaxDSW}}(\mu,\nu)\big|
   \;\leq\;
   d_{\text{MaxDSW}}(\hat{\mu}_n,\mu)
   \,+\,
   d_{\text{MaxDSW}}(\hat{\nu}_n,\nu).
\]
Therefore, the corresponding sample complexity is controlled by applying the previous estimate separately to \(\mu\) and \(\nu\), with the tolerance replaced by \(\varepsilon/2\).

\section{Experiments and applications}\label{sec:numericalconsiderations}
\noindent In this section, we present and discuss several applications of the empirical Max-D-SW distance.
First, we investigate through simulations how the Max-D-SW distance behaves relative to the classical Wasserstein distance and several of its variants discussed in this paper in settings where multiple sample properties vary simultaneously.
Next, we illustrate how its discriminative power can be used to visualize samples from distinct but related distributions in a meaningful way, via low-dimensional embeddings obtained with MDS based on their Max-D-SW distances. We further analyze how these visualizations vary with sample size. In particular, for the smaller sample sizes considered, Max-D-SW provides a better characterization of the underlying structure.
Finally, the section concludes with a concrete application introduced by Sutherland  \textit{et al.} \cite{sutherlandsamples}, in which one seeks meaningful representations of pictures of objects under different translations, rotations and observed by sets of feature vectors.

\subsection{Distance comparison against a reference measure}
In this case, we examine how a distance measure $d_{test}$ behaves between a reference distribution and a family of distinct but related distributions represented through their respective samples. We take for $d_{test}$ the Max-D-SW distance, the Wasserstein distance and the Max-Sliced Wasserstein distance. We compare their behavior under a simple, artificially constructed transformation of measures.\\   
 
\noindent The specific context is formulated as follows: consider a bivariate standard Gaussian distribution $\mu$ with mean zero and identity covariance matrix that will be playing the role of reference measure. Consider as well a collection of Gaussian distributions $\{\nu_i\}_{i=1}^m$ with $\nu_i$ having mean $[0,1]$ and covariance matrix $\Sigma_i$, with 
$$\Sigma_i = \begin{bmatrix}
		1 & s_i \\
		s_i & 1 
	\end{bmatrix},$$ 
for some $0<|s_i|<1$. The key point is that we do not have direct access to the $\nu_i$; instead we only observe samples drawn from them. Since we are artificially constructing the measures to test effectiveness of performance, we have the freedom of choosing the $s_i$ in such a way that we have some geometric intuition about them. To this end, we will impose the condition that the $s_i'$s being such that $|s_j|>|s_i|$ when  $j>i$. Given this setting, one would expect that the distance will be increasing with the index $i$.\\

\begin{figure}
	  	  	\centering
	  	  	\includegraphics[width=0.6\linewidth]{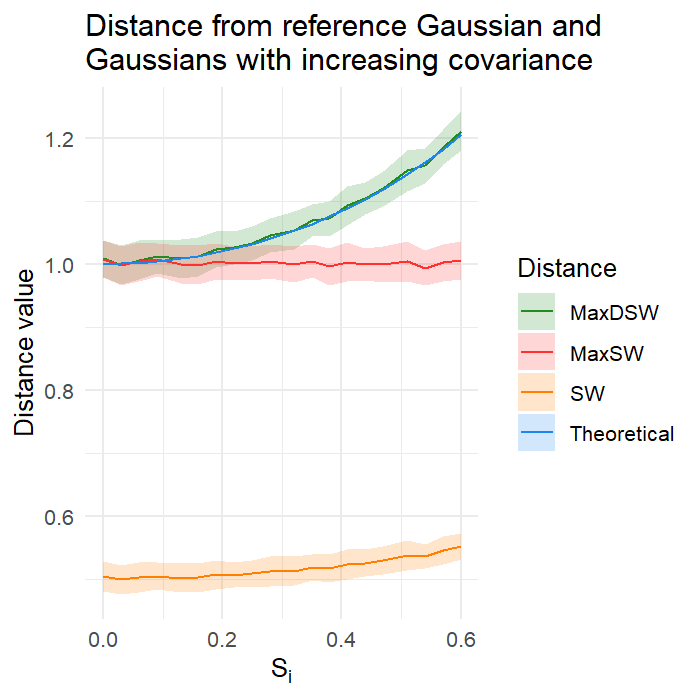}
	  	  	\caption{Mean and variance of the distances between the reference Gaussian distribution $\mu$ and Gaussian distributions with increasing covariance parameter $S_i$. }
	  	  	\label{fig:screenshot001}
	  	  \end{figure}

\noindent The setting is simple enough to derive an explicit expression for the projected Wasserstein distance in the case \(p=2\). To this end, consider a projection direction \(\theta^\intercal=(\theta_1,\theta_2)\in\mathbb S^1\). The projected law of \(\mu\) is
\[
(R^\theta)_\#\mu=N(0,1),
\]
whereas the projected law of \(\nu_i\) is
\[
(R^\theta)_\#\nu_i
=
N\left(\theta_2,\theta^\intercal\Sigma_i\theta\right)
=
N\left(\theta_2,1+2s_i\theta_1\theta_2\right).
\]
Therefore, using the explicit expression for the \(W_2\)-distance between one-dimensional Gaussian laws, we obtain
\begin{equation}\label{w}
W_2^2\big((R^\theta)_\#\mu,(R^\theta)_\#\nu_i\big)
=
\theta_2^2+
\left(1-\sqrt{1+2s_i\theta_1\theta_2}\right)^2.
\end{equation}
To compute the Max-Sliced Wasserstein distance, we maximize \eqref{w} over \(\theta\in\mathbb S^1\). In the range \(|s_i|\leq 1/2\), this maximum is equal to \(1\). Indeed,
\[
\left|1-\sqrt{1+2s_i\theta_1\theta_2}\right|
=
\frac{|2s_i\theta_1\theta_2|}
     {1+\sqrt{1+2s_i\theta_1\theta_2}}
\leq |\theta_1|.
\]
Consequently,
\[
W_2^2\big((R^\theta)_\#\mu,(R^\theta)_\#\nu_i\big)
\leq
\theta_2^2+\theta_1^2
=
1.
\]
Since equality is attained at \(\theta=(0,1)\), the Max-Sliced Wasserstein distance remains equal to \(1\) throughout this range of values of \(s_i\). This is a troublesome situation for the purpose of this example, because the different distributions \(\nu_i\) are not distinguished by Max-SW even though their covariance matrices are different. The reason is that the maximizing projection is dominated by the displacement of the mean, while the variation in the covariance parameter is ignored. In simpler terms, one projection direction is not enough to detect the additional motion. As one might expect, the Max-D-Sliced Wasserstein distance captures these differences effectively, producing a map \( i \mapsto d_{\text{MaxDSW}}(\nu_i, \mu) \) whose level sets are numerically well-separated. This behavior is clearly illustrated in Figure~\ref{fig:screenshot001}, which shows a pronounced and consistent trend, in notorious contrast to the Max-Sliced  Wasserstein distance.\\ 
	  
	  	  \noindent Finally, we include the corresponding analysis for the case of the Sliced Wasserstein distance and Max-D-SW distance, computed numerically. Regardless of the fact that we have an explicit expression for the Max-Sliced distance, to even up the comparison between all test distances, we used as well a numerical scheme for estimating from random samples. Also, given that the distribution of the samples is Gaussian, there exists a closed expression for the Wasserstein distance between the samples, which we identify by ``theoretical distance''. All the other distances, were computed numerically over samples of size $n=10,000$ from the Gaussian distributions $\mu$ and $\{\nu_i\}$ with $s \in [0, 0.6]$ , with $l=600$ projection directions and 50 repetitions of the experiment. In Figure \ref{fig:screenshot001}, the mean of the distances obtained by the 50 repetitions of the experiment is presented by a continuous line and the ribbon represents the standard deviation from the mean of this distances.\\ 	  	  
	  	  
	  	  \noindent As expected, Max-Sliced Wasserstein fails to distinguish between the different distributions and assigns the same distance between $\mu$ and every $\nu_i$. In this case, Sliced Wasserstein discriminates correctly between the different $\nu_i$'s but is not as discriminative as the Max-D-SW, which in this case closely follows the theoretical distance.\\

\subsection{Max-D-SW for pattern detection with heavy tailed distributions}
Next, we explore the behaviour of the Sliced Wasserstein, Max Sliced Wasserstein and Max-D-Sliced Wasserstein distances for p=1 and p=2 in a dimensionality reduction problem. The objective of the experiment taken from  \cite{sutherlandsamples} is to embed a collection of 2 dimensional samples of Gaussian distributions into a 2 dimensional space, that is, to represent each distribution by a point in $\mathbb{R}^2$. In our case, instead of considering locally linear embedding (as in \cite{sutherlandsamples}), we utilize multidimensional scaling as our main tool for dimensionality reduction. For Gaussian distributions, the results obtained by the different distances is similar even when considering small sample sizes. Therefore, in addition to the Gaussian case, we consider lognormal distributions in order to assess the behaviour of the methods when samples with a long tail are present.\\

\noindent The specific configuration is as follows: consider 40 centered lognormal distributions in dimension 2, where the entries of the covariance matrices are rotated versions by some angle $\alpha_i$ of an original covariance matrix. For each $i \in \{1,\dots,40\}$, the covariance matrix is:
 \begin{eqnarray*}
\Sigma_i  &=& R(\alpha_i)\left(\begin{array}{cc}
		3.3 & 0\\
		0 & 0.2
	\end{array}\right) R(\alpha_i)^\intercal 	
\end{eqnarray*}	
and $R(\alpha_i)$ corresponds to the 2D rotation matrix with rotation angle $\alpha_i = (i-1)/12.7324$.
The difference between each sample is given by the rotation angle $\alpha_i \in \{ 0, ...,40\}/12.7324$. The samples generated have an order given by the index.\\

 \noindent For the estimation of the three distances, several numbers of projection directions were considered, up to $l=1200$. Since increasing the number of projection directions did not yield considerably different results, we present results based on $l=200$ projection directions.\\

\noindent Different sample sizes were considered, and the results of MDS for MaxSW, SW, and Max-D-SW with $p=1$ are presented in Table \ref{tabla1}. Since a very large sample size ($n=100,000$) was used for the last row, we regard these figures as a proxy for the ``real pattern'' that would be obtained with infinitely many samples and projection directions. The configurations obtained from multidimensional scaling (MDS) were isotropically normalized and subsequently aligned using Procrustes analysis. This procedure removes indeterminacies due to translation, rotation, reflection, and global scaling, thereby placing the configurations in a common coordinate system and enabling a meaningful geometric comparison across distance matrices. Note that despite the different vertical orientation, the ``real pattern'' figures correspond to the same data, as the sample ID order is preserved.\\

	\begin{table}[h!]
		\centering
		\begin{tabular}{|c|c|c|c|}
			\hline
			& \textbf{SW} & \textbf{MaxSW} & \textbf{Max-D-SW}  \\
			\hline
			\textbf{n=200} &
			\includegraphics[width=0.18\textwidth]{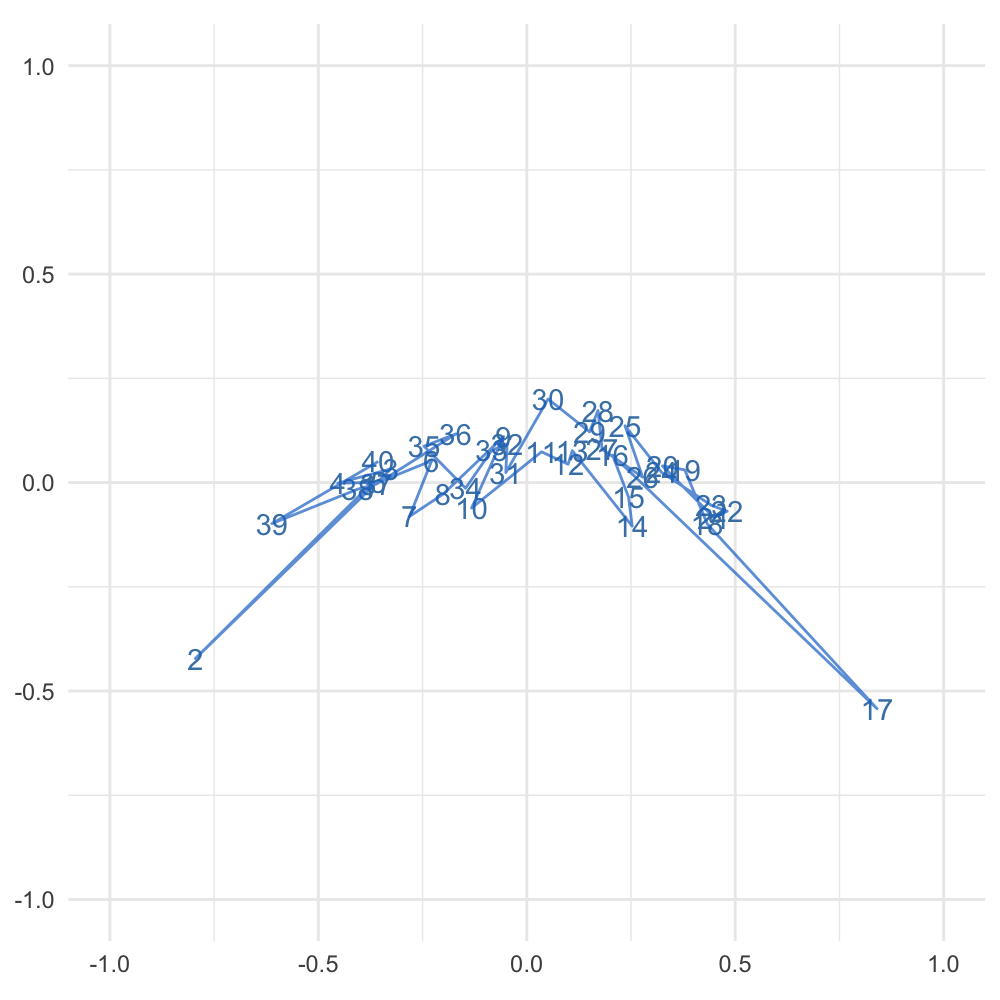} &
			\includegraphics[width=0.18\textwidth]{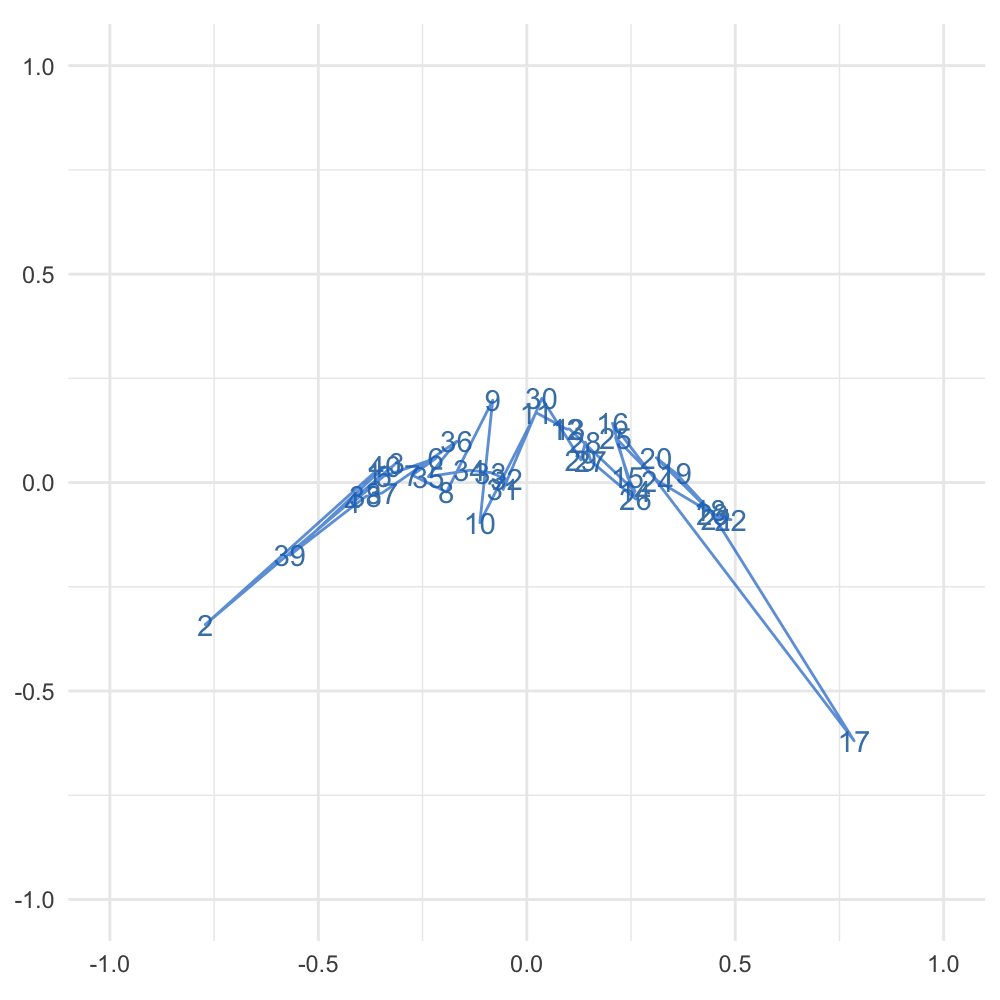} &
			\includegraphics[width=0.18\textwidth]{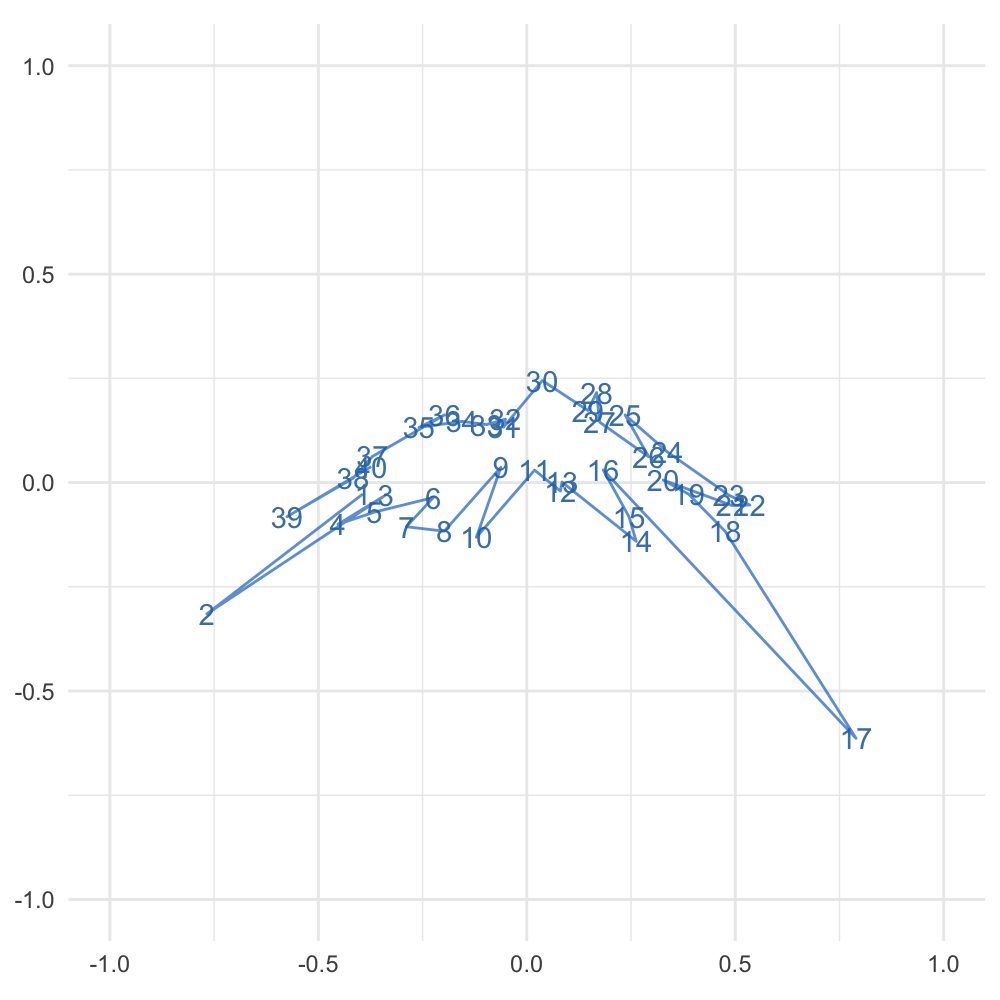} \\
			\hline
			\textbf{n=500} &
			\includegraphics[width=0.18\textwidth]{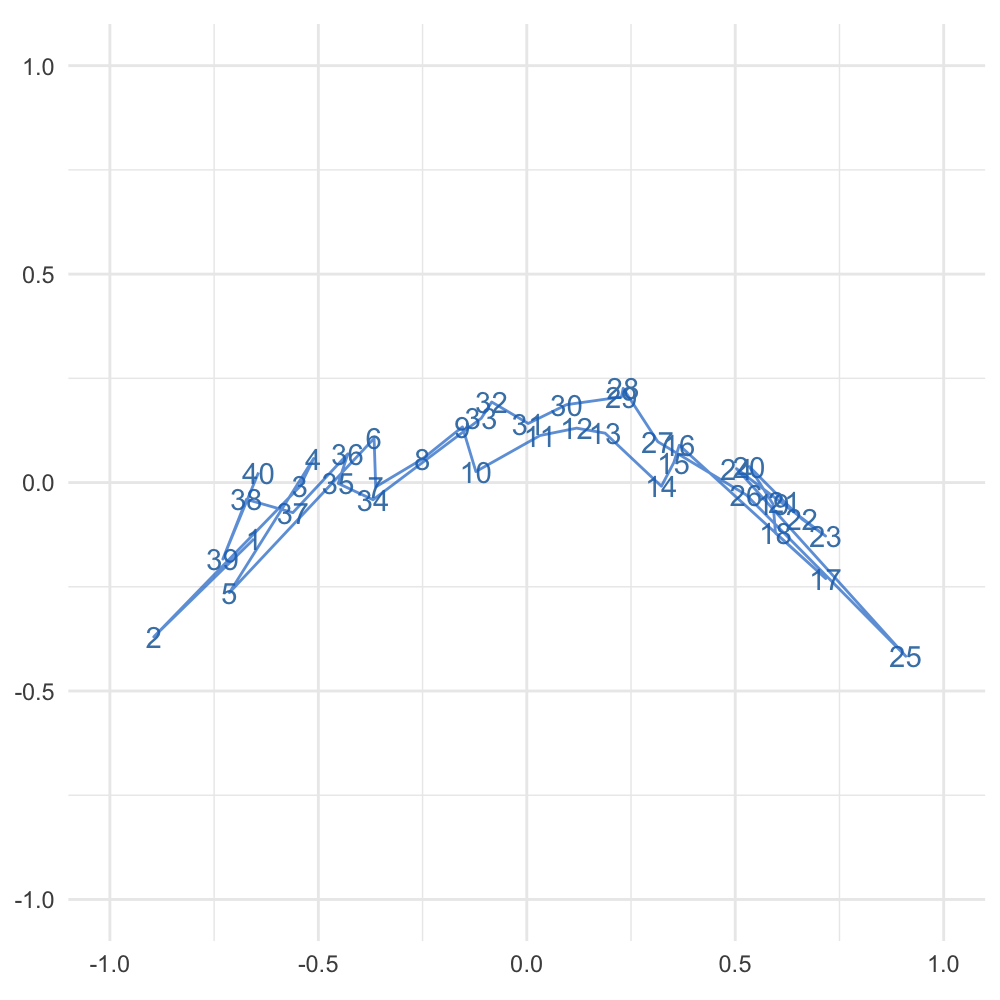} &
			\includegraphics[width=0.18\textwidth]{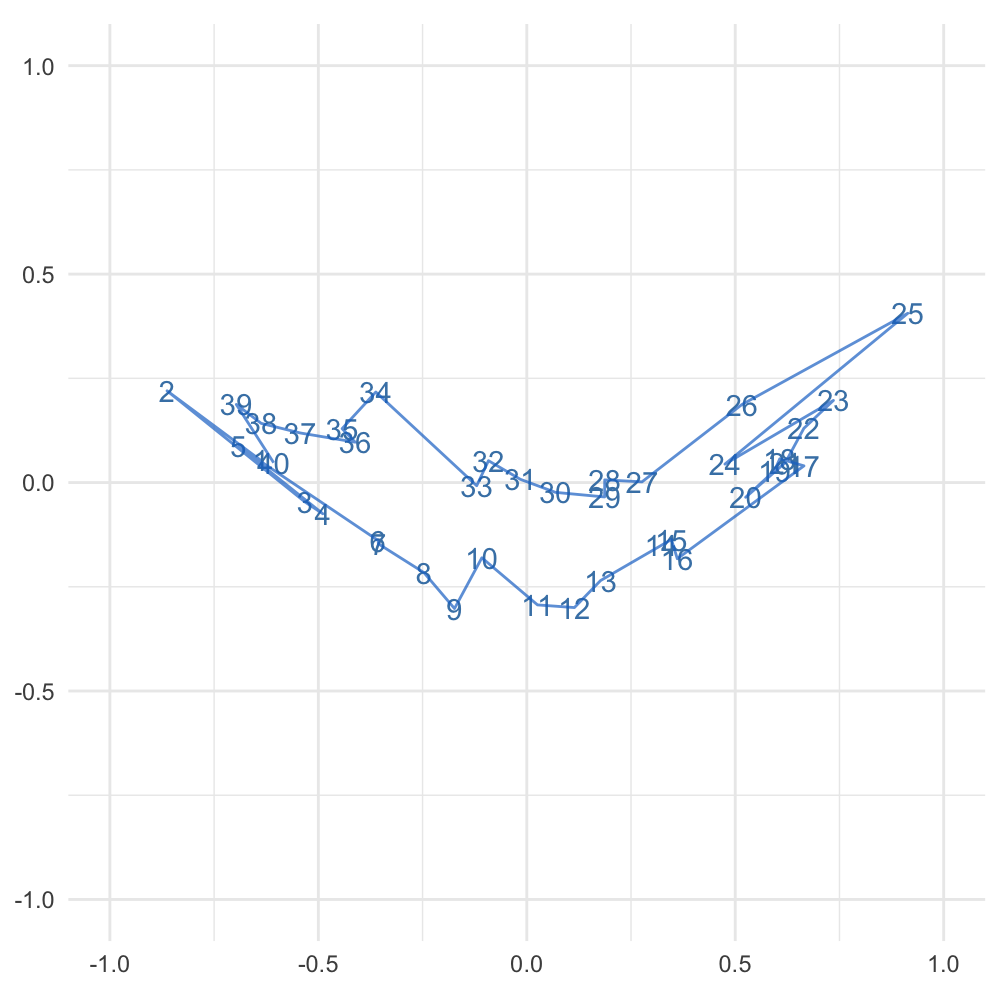} &
			\includegraphics[width=0.18\textwidth]{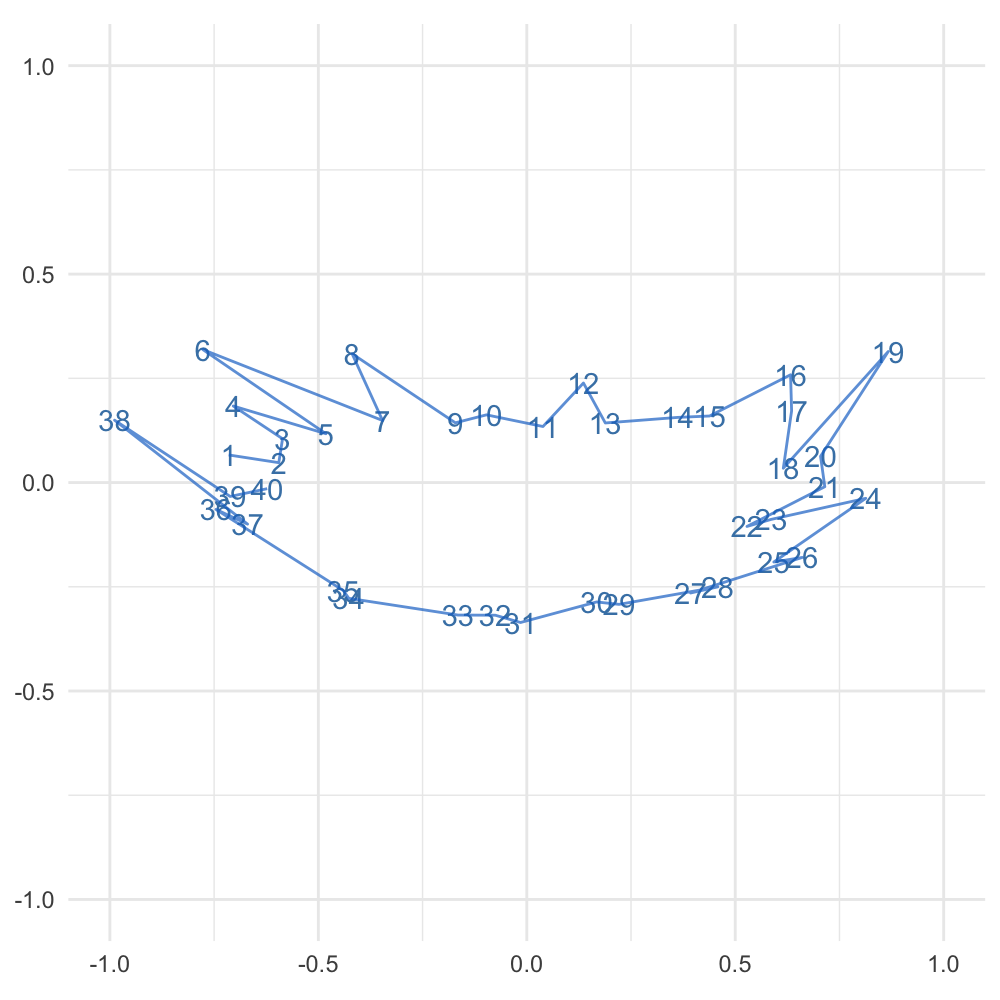} \\
			\hline
			\textbf{n=1000} &
			\includegraphics[width=0.18\textwidth]{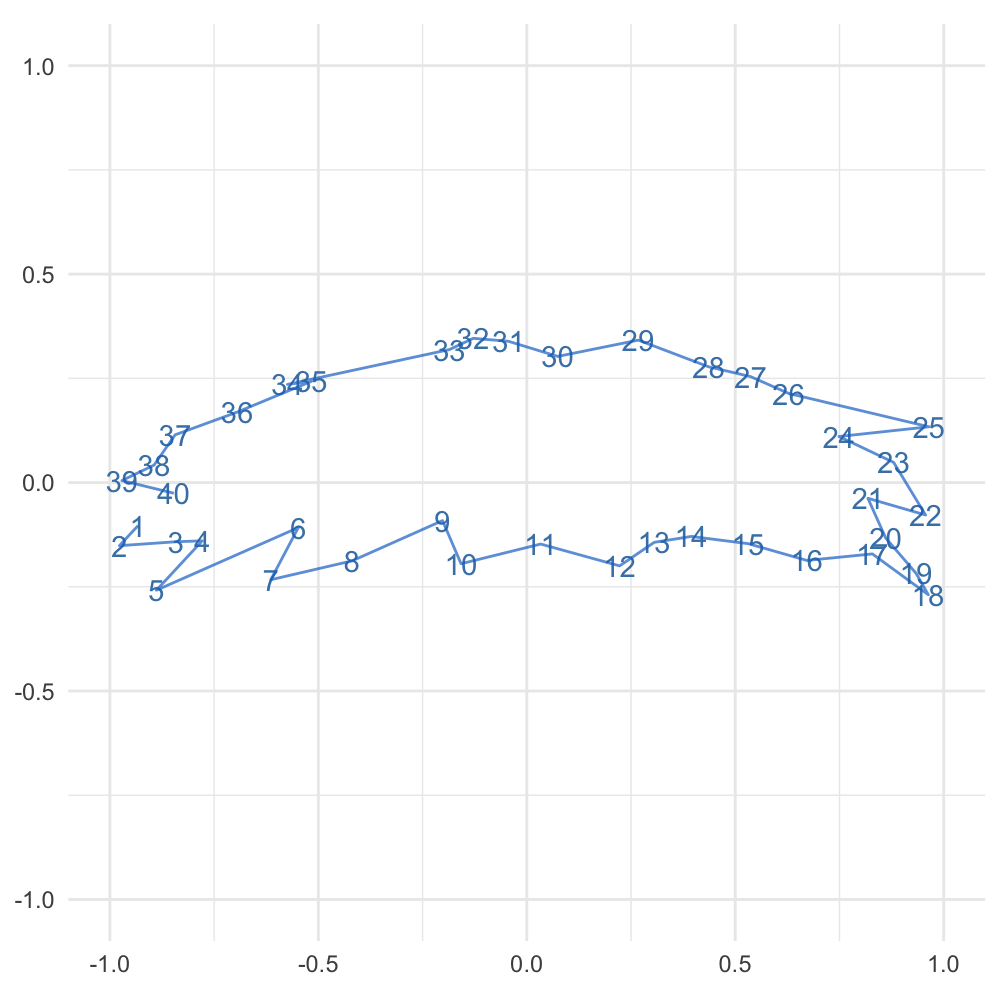} &
			\includegraphics[width=0.18\textwidth]{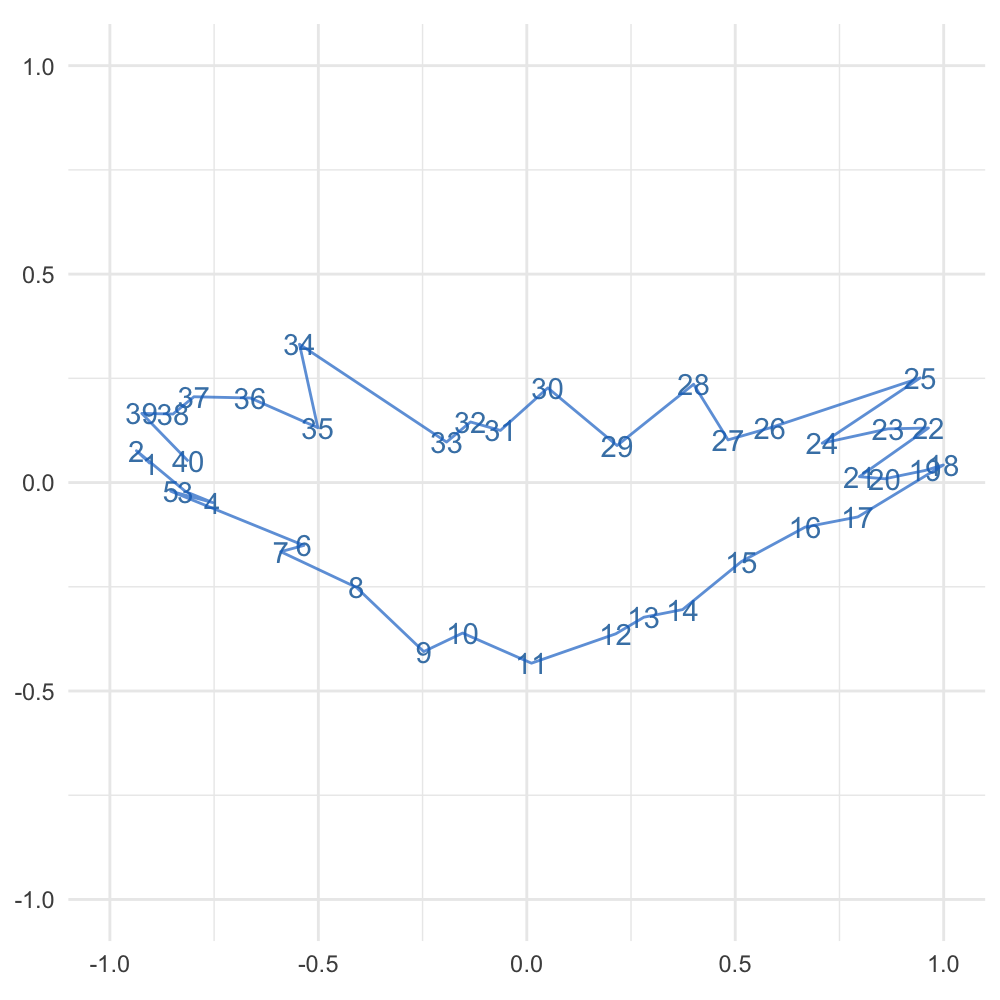} &
			\includegraphics[width=0.18\textwidth]{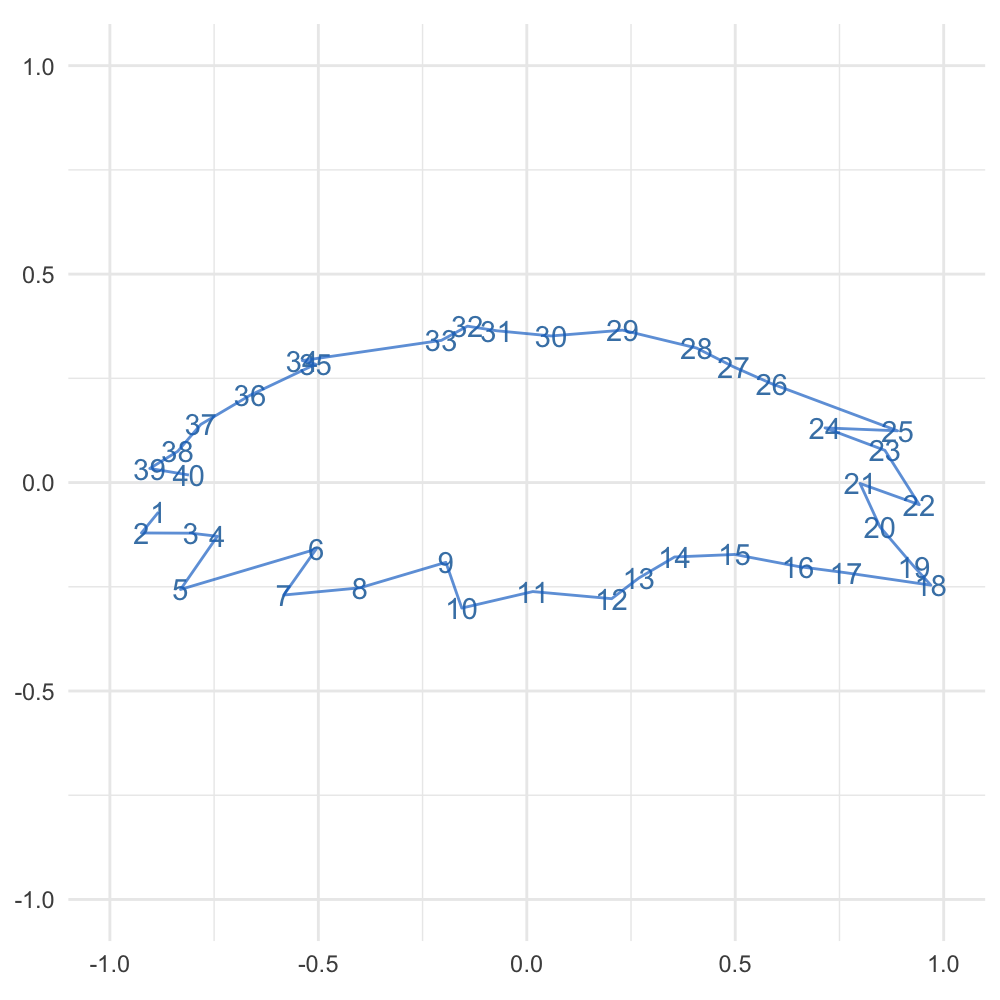} \\
			\hline
			\textbf{n=2500} &
			\includegraphics[width=0.18\textwidth]{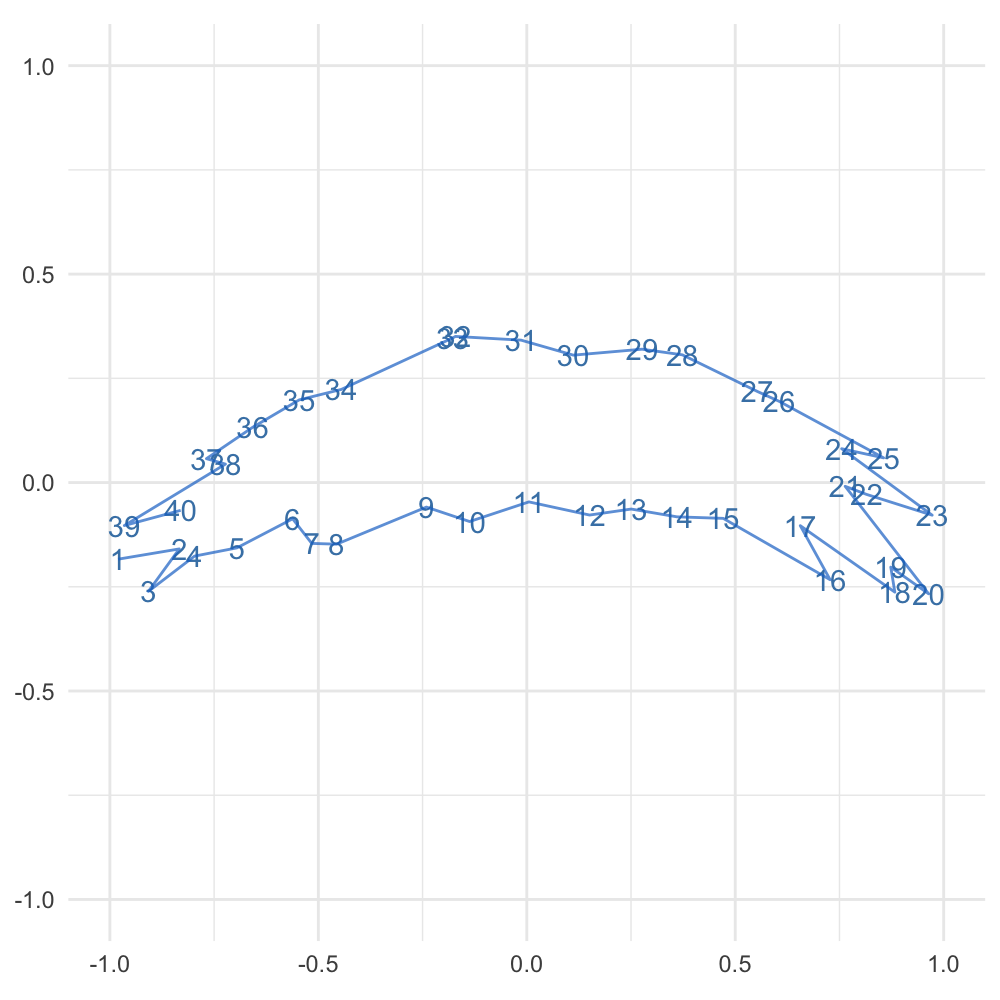} &
			\includegraphics[width=0.18\textwidth]{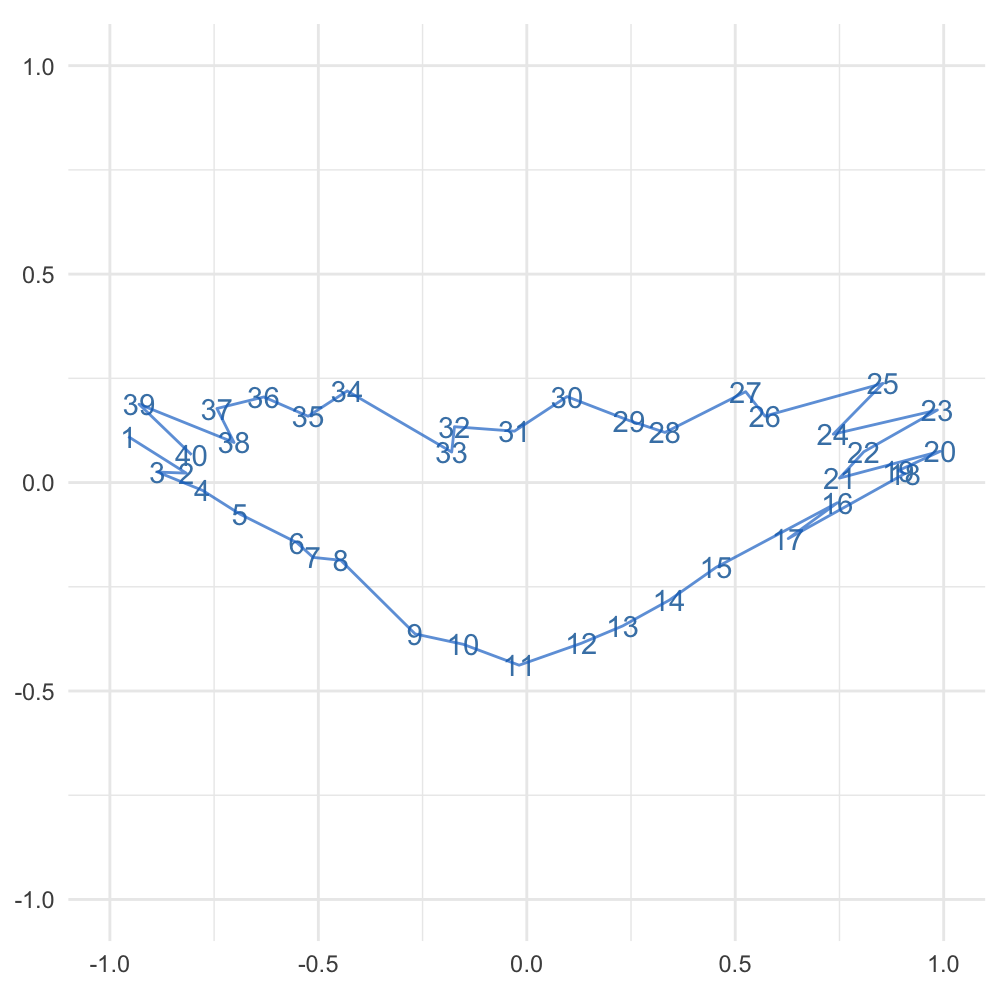} &
			\includegraphics[width=0.18\textwidth]{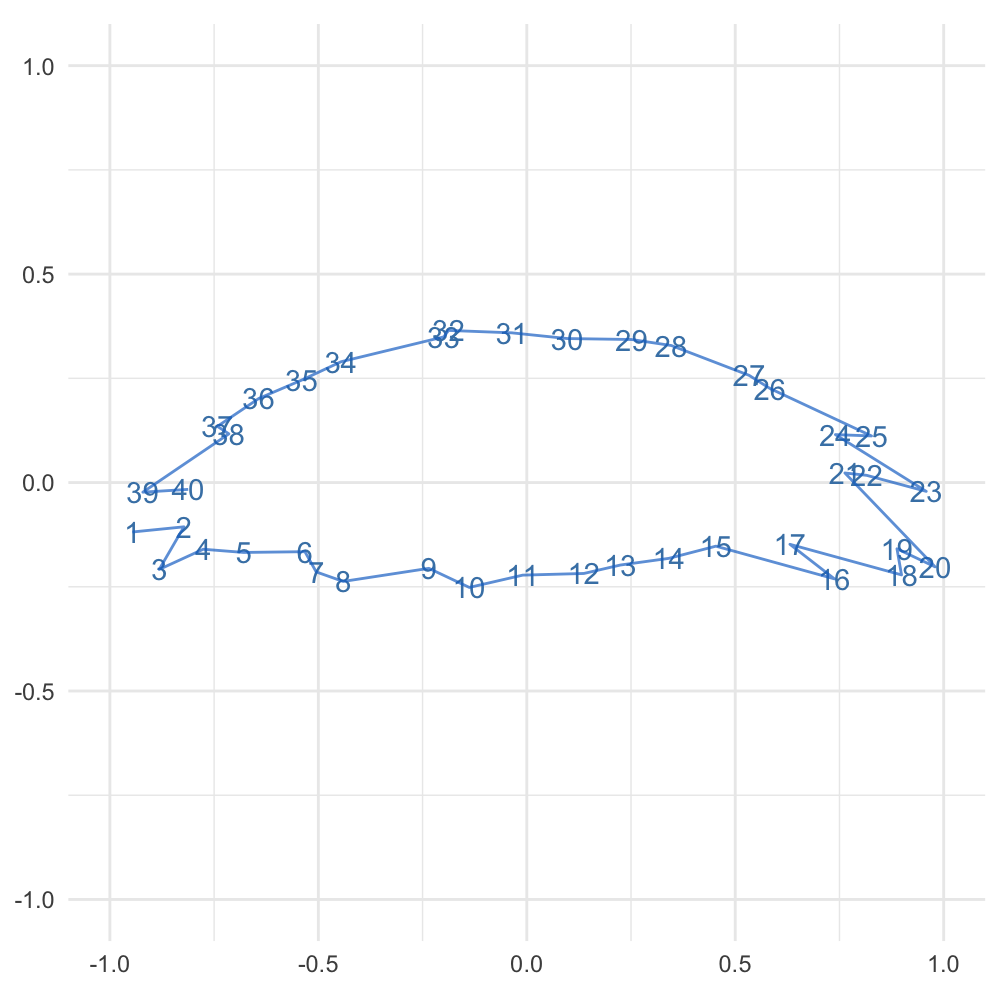}  \\
			\hline
			 \textbf{n=100000} &
			\includegraphics[width=0.18\textwidth]{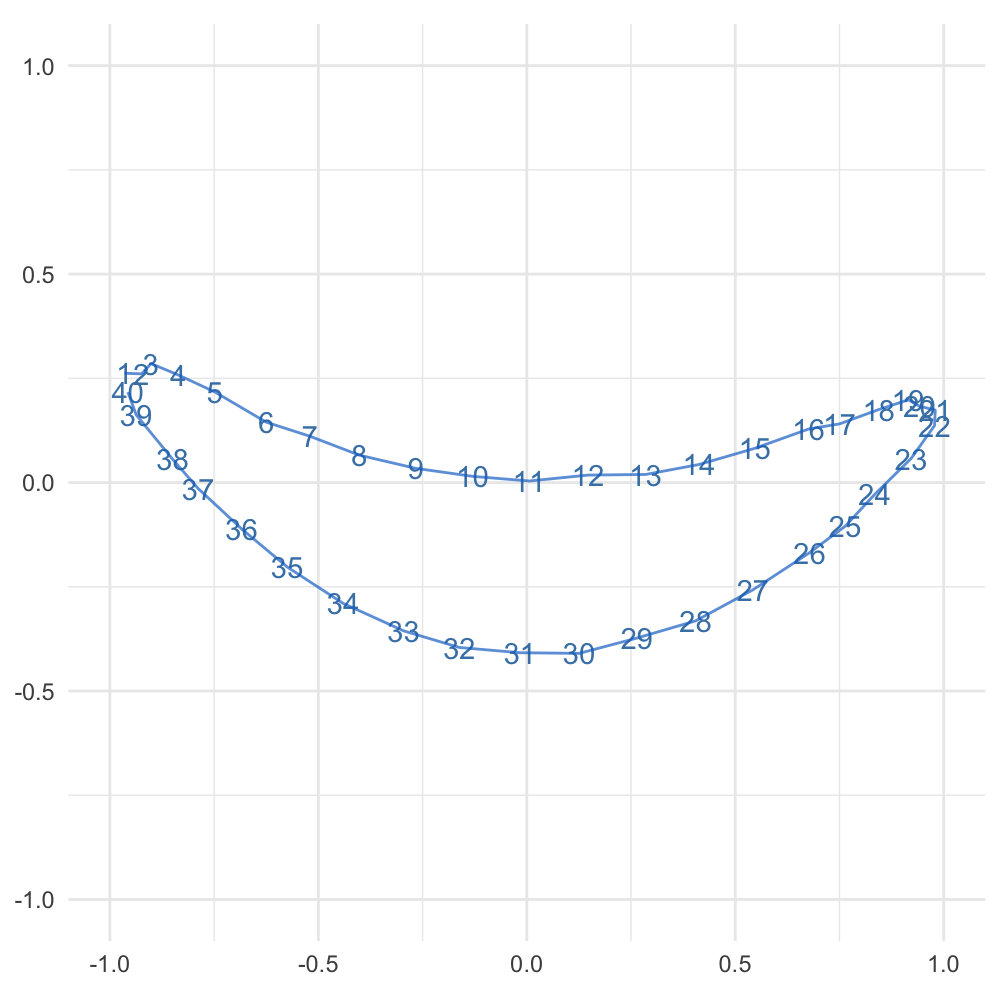} &
			\includegraphics[width=0.18\textwidth]{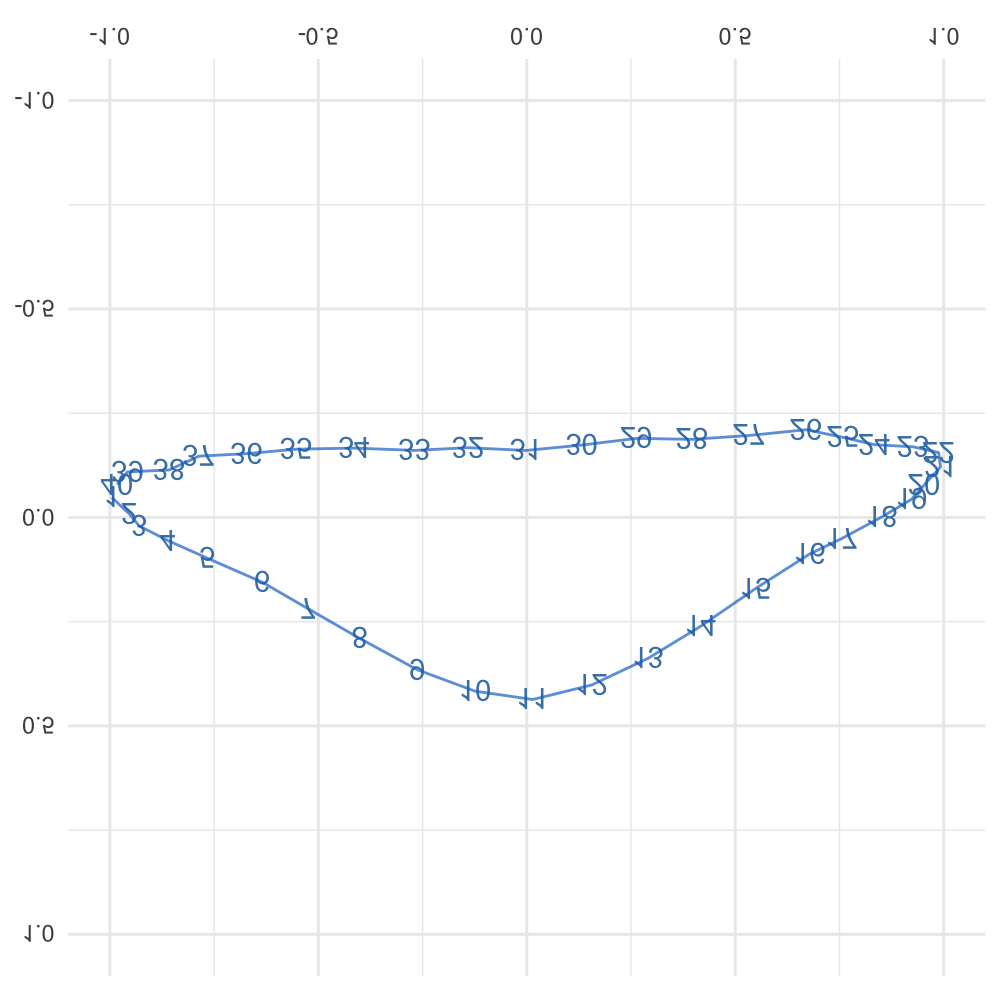} &
			\includegraphics[width=0.18\textwidth]{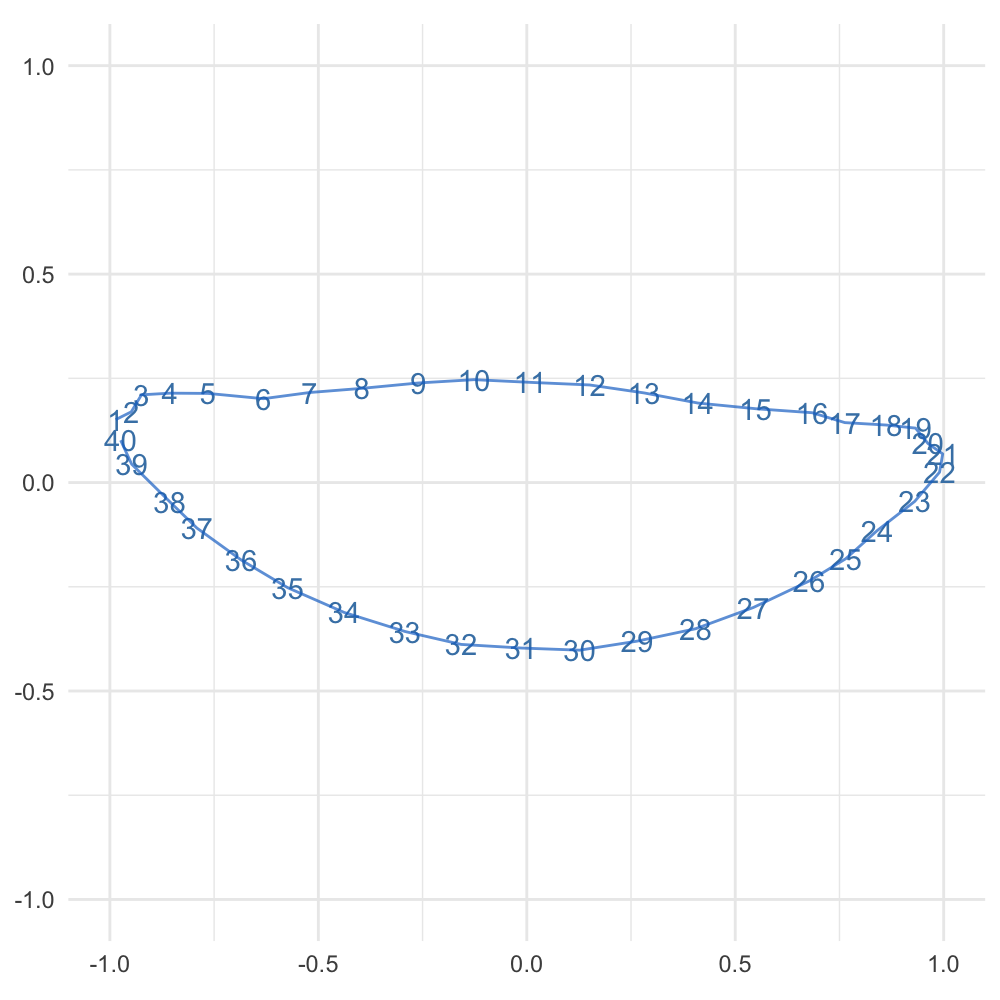}  \\
			\hline
		\end{tabular}
		\caption{Patterns obtained by MDS with different sample sizes for SW, Max-SW, and Max-D-SW with $p=1$ and $l=200$ projection directions.}
		\label{tabla1}
	\end{table}

\noindent In all cases, the estimation of the underlying pattern improves as the sample size increases.  Max-D-SW produces a configuration closer to the ``real pattern'' than the other two distances, especially for small sample sizes.\\

\noindent We now consider the same experiment with $p=2$. In this case, all the distances fail to give a clear figure when considering lognormal distributions with the parameters mentioned above, but if the heavy tail of the distribution is reduced, some patterns can be seen. This suggests that, when working with heavy-tailed distributions, using $p=1$ may be preferable.    The parameters of the lognormal distribution for the case $p=2$ are:
 \begin{eqnarray*}
	\Sigma_i  &=& R(\alpha_i)\left(\begin{array}{cc}
		1 & 0\\
		0 & 0.2
	\end{array}\right) R(\alpha_i)^\intercal 	
\end{eqnarray*}	
and $R(\alpha_i)$ corresponds to the 2D rotation matrix with rotation angle $\alpha_i = (i-1)/12.7324$.\\

\begin{table}[h!]
	\centering
	\begin{tabular}{|c|c|c|c|}
		\hline
		& \textbf{SW} & \textbf{MaxSW} & \textbf{Max-D-SW}  \\
				\hline
		\textbf{n=500} &
		\includegraphics[width=0.18\textwidth]{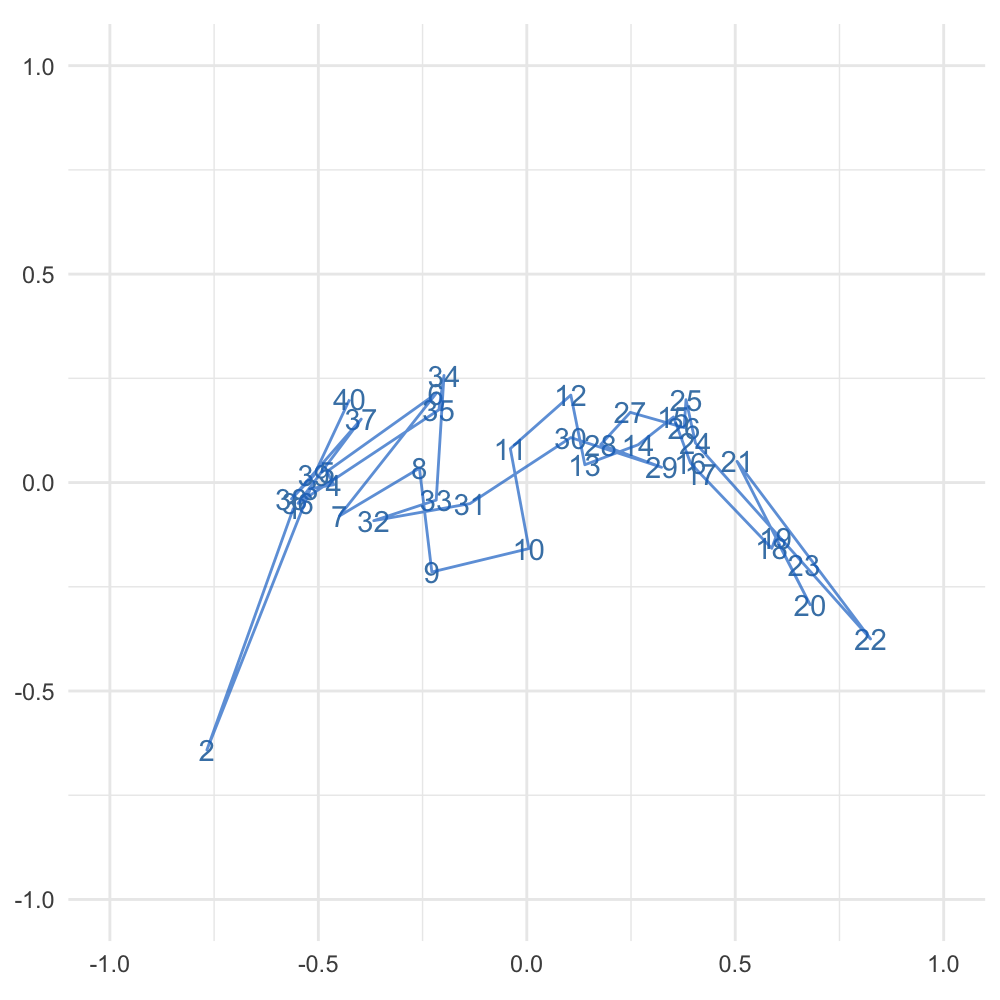} &
		\includegraphics[width=0.18\textwidth]{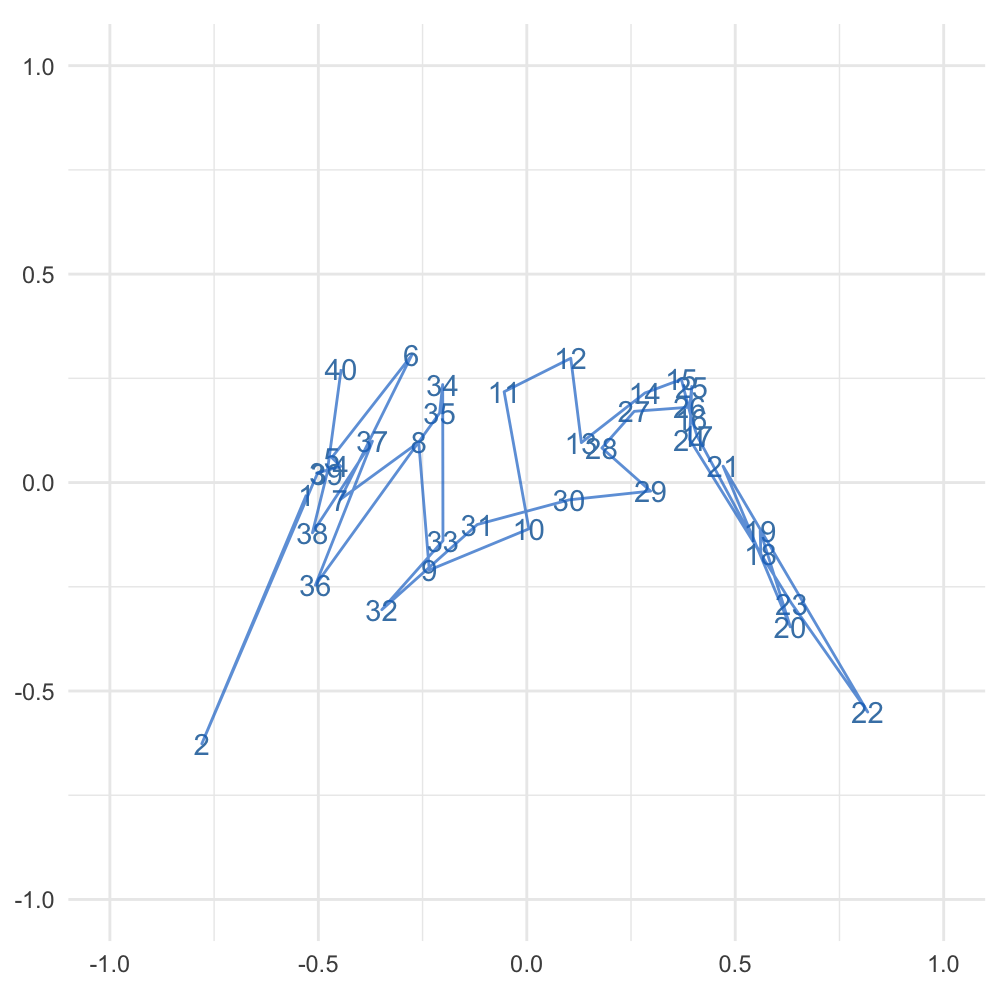} &
		\includegraphics[width=0.18\textwidth]{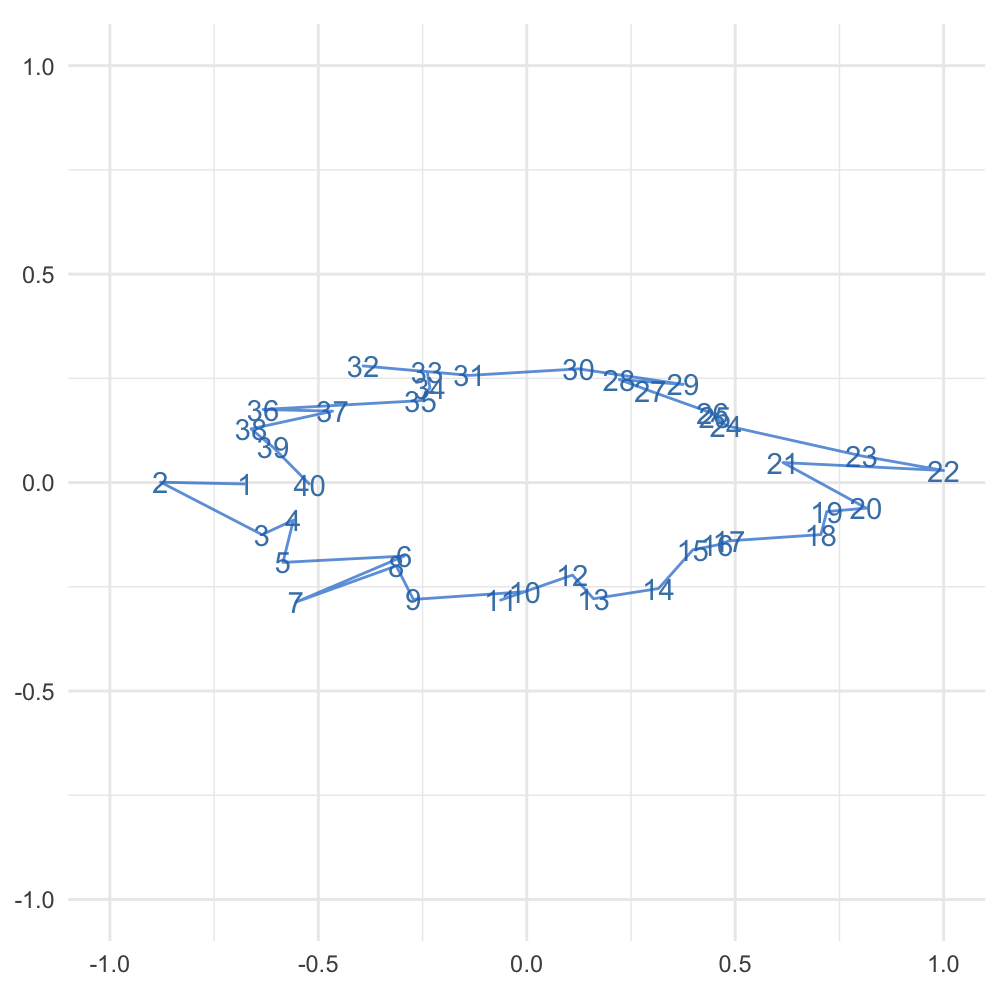} \\
		\hline
		\textbf{n=1000} &
		\includegraphics[width=0.18\textwidth]{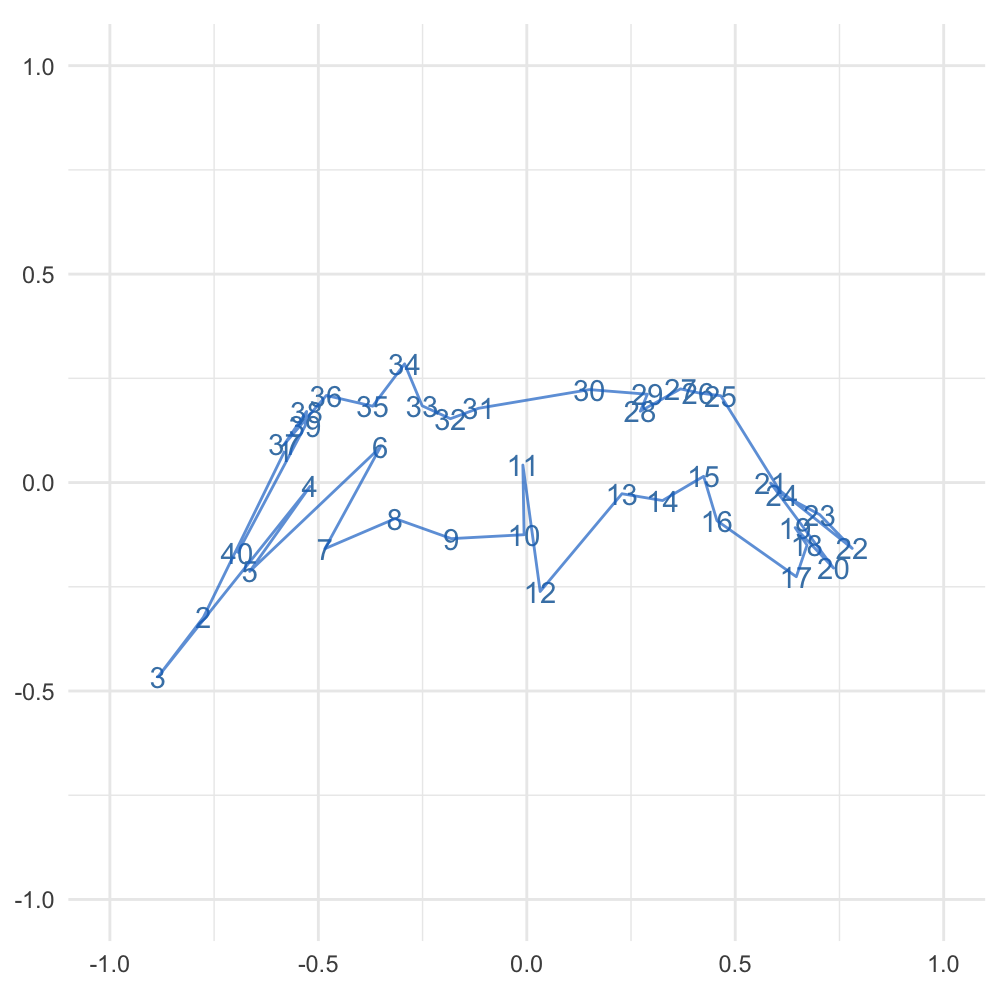} &
		\includegraphics[width=0.18\textwidth]{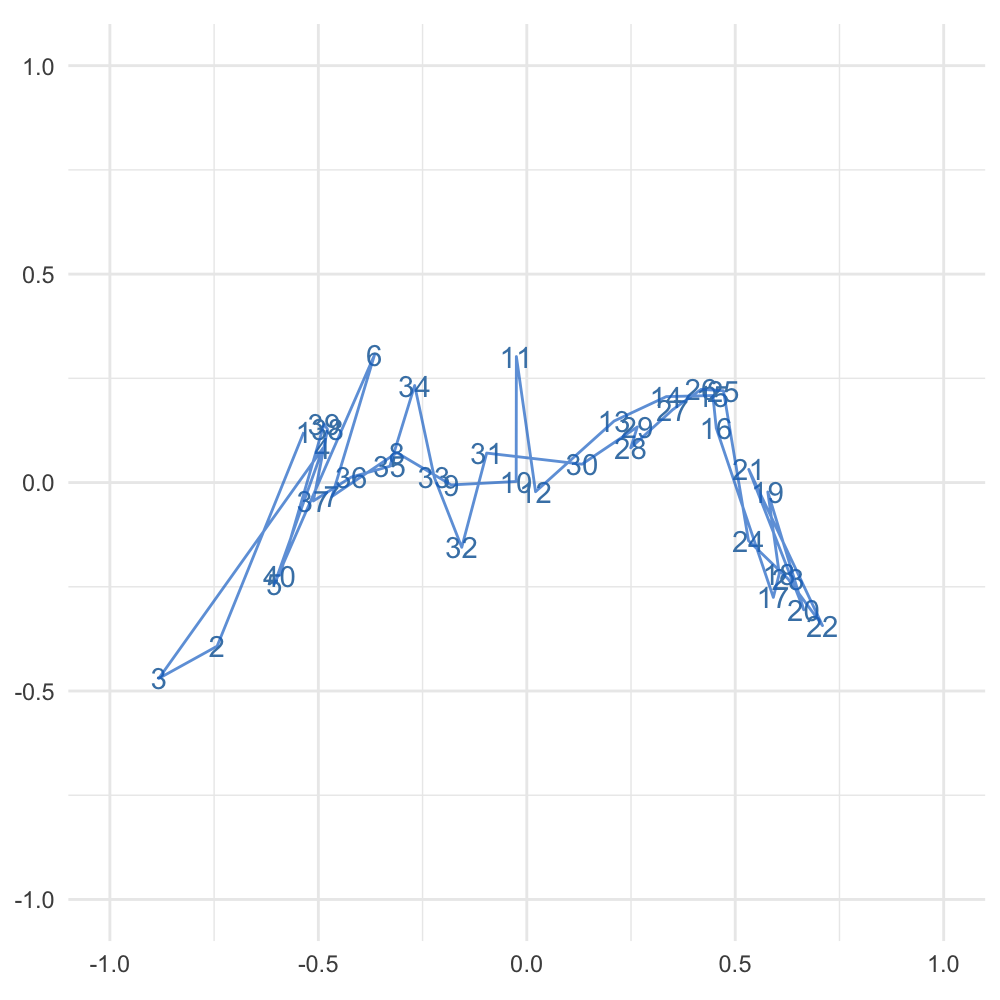} &
		\includegraphics[width=0.18\textwidth]{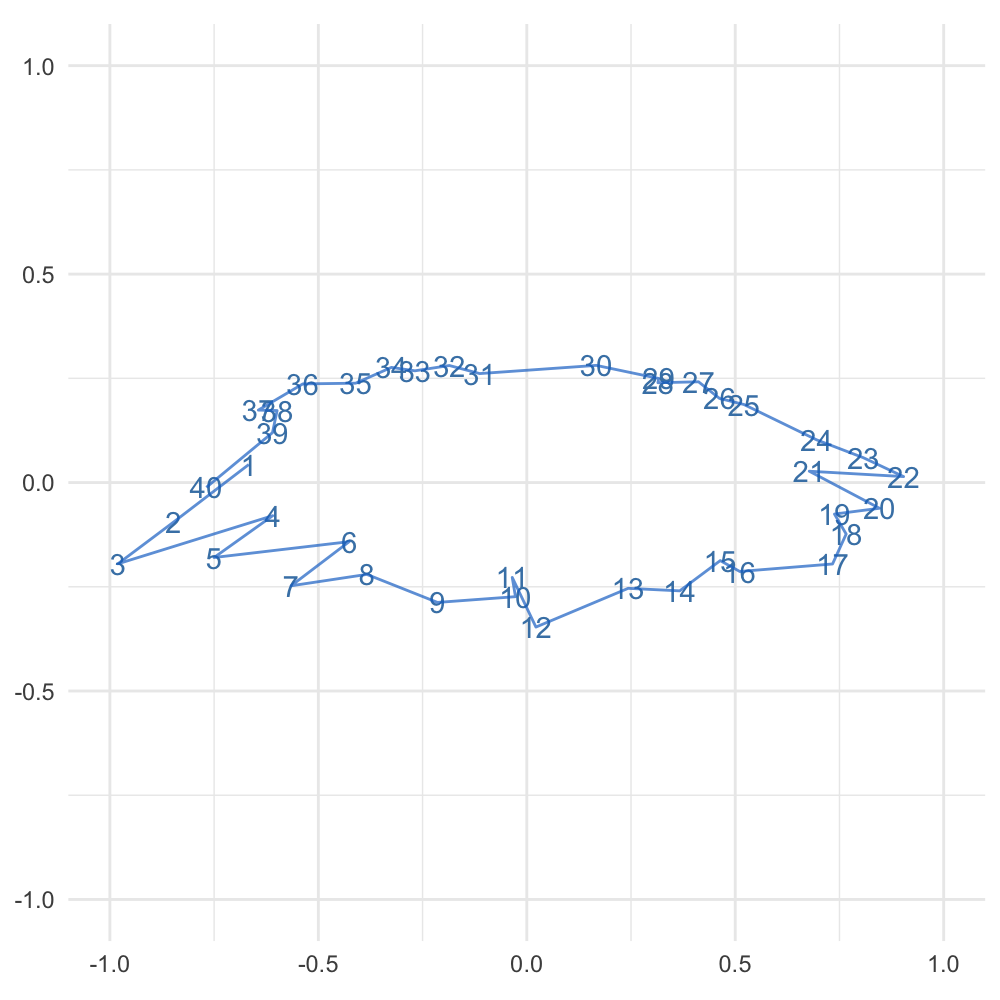} \\
		\hline
		\textbf{n=2500} &
		\includegraphics[width=0.18\textwidth]{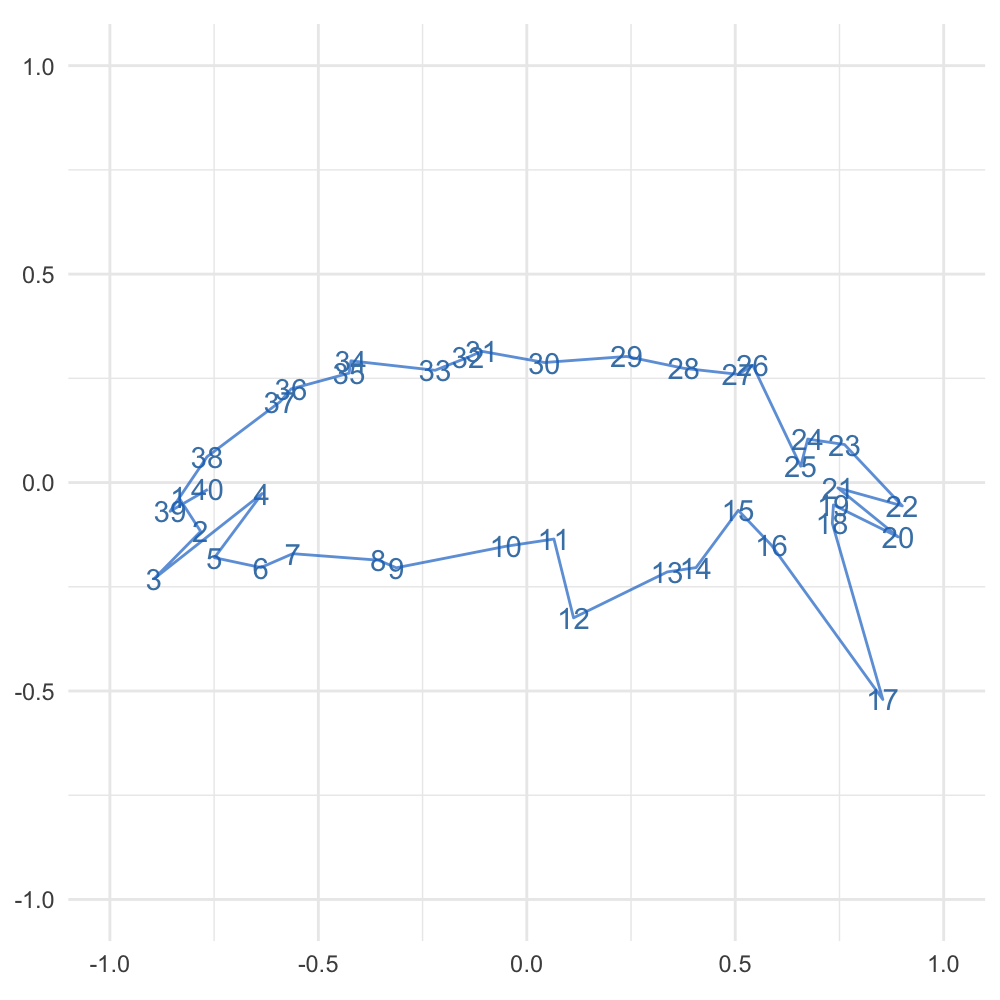} &
		\includegraphics[width=0.18\textwidth]{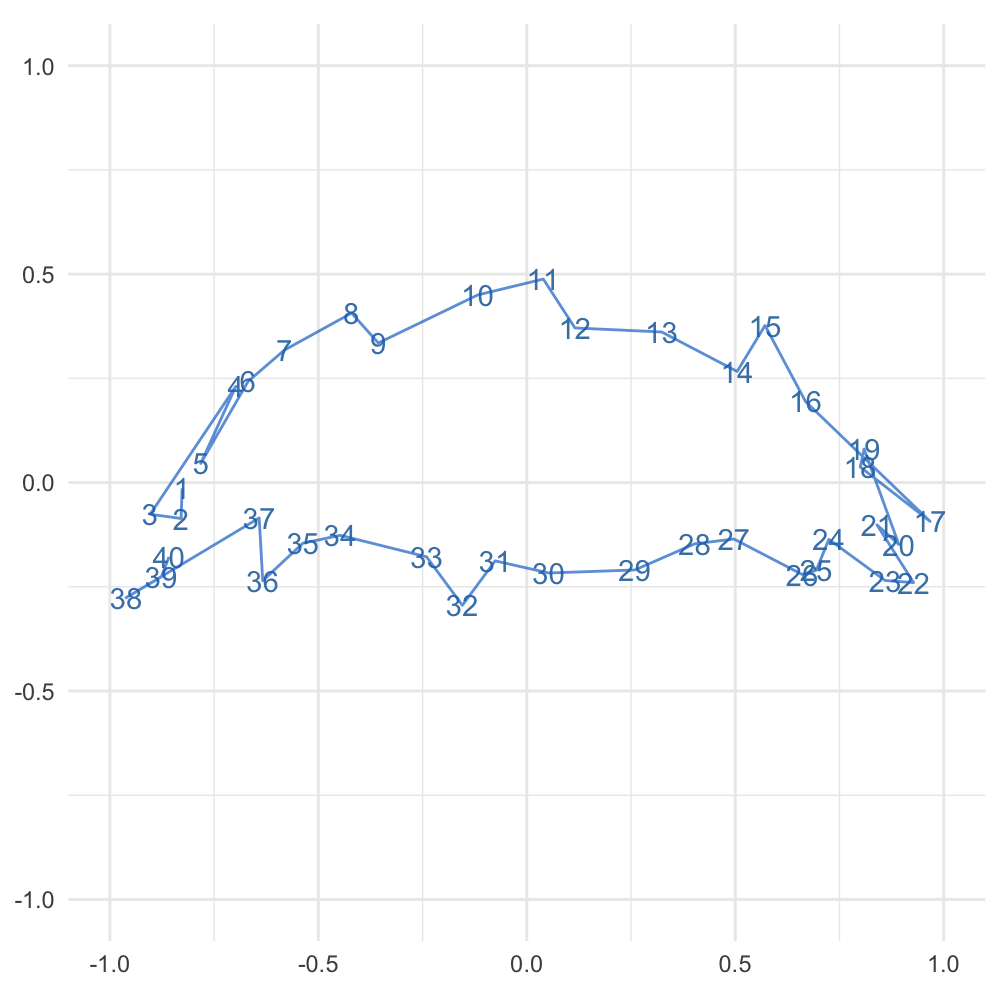} &
		\includegraphics[width=0.18\textwidth]{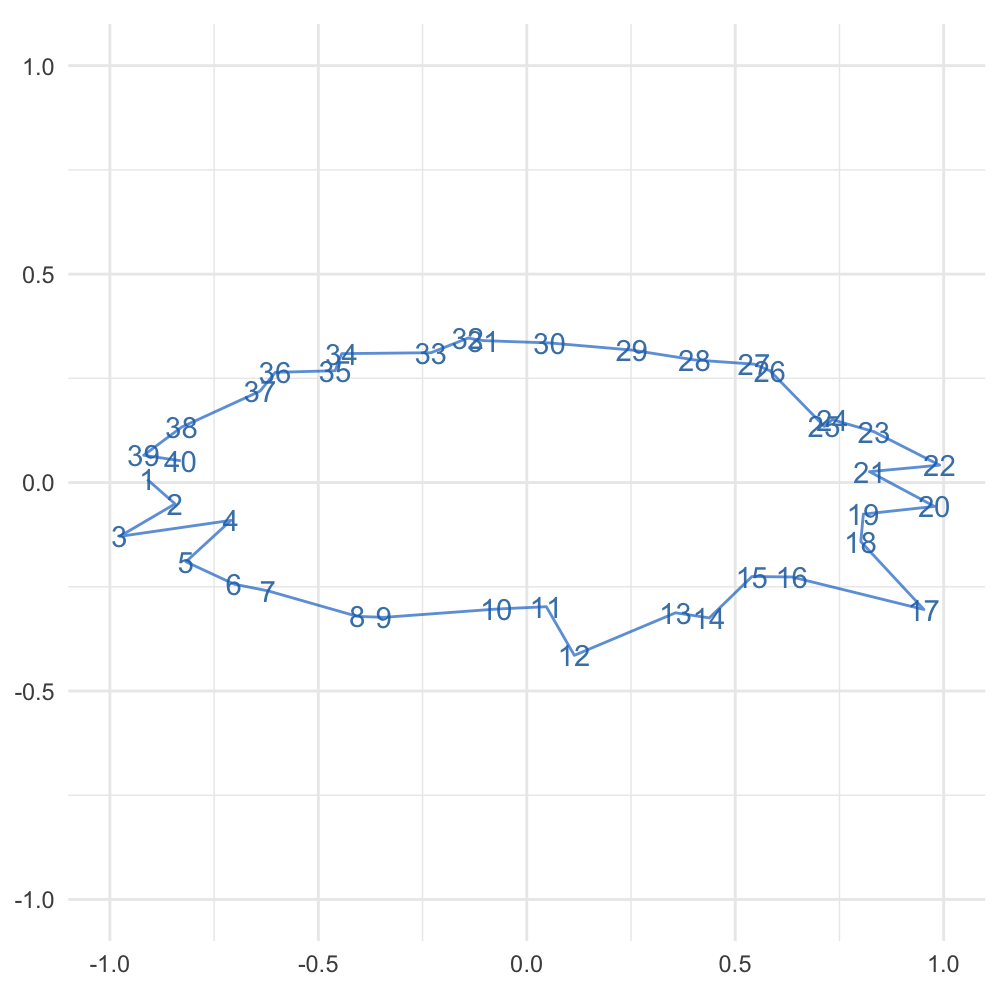}  \\
		\hline
		\textbf{n=100000} &
		\includegraphics[width=0.18\textwidth]{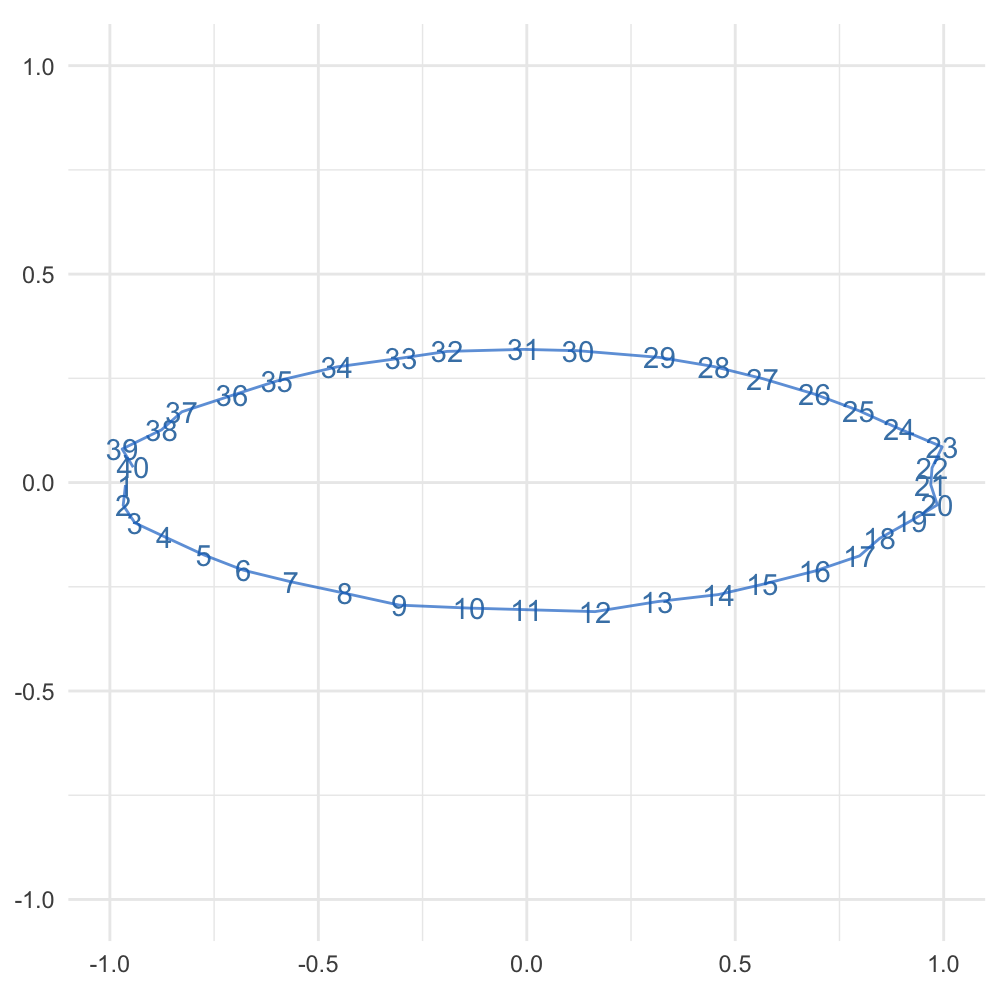} &
		\includegraphics[width=0.18\textwidth]{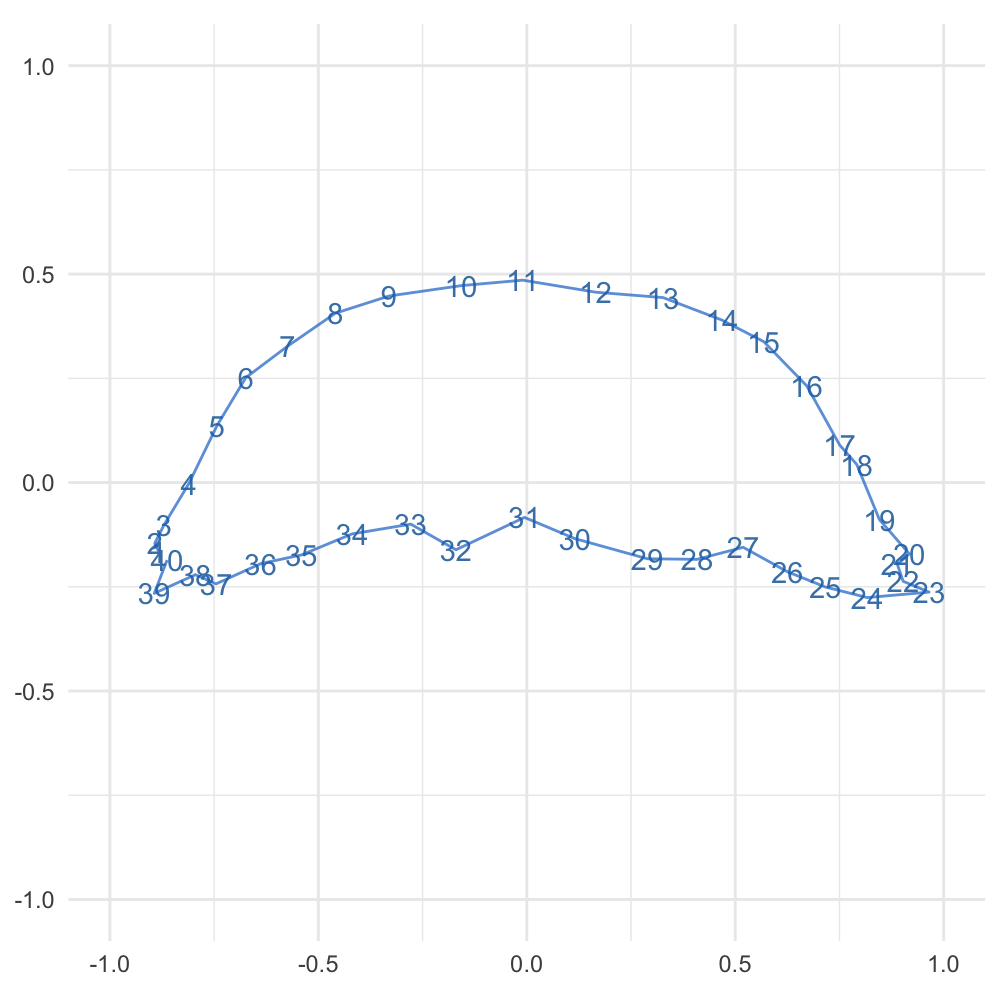} &
		\includegraphics[width=0.18\textwidth]{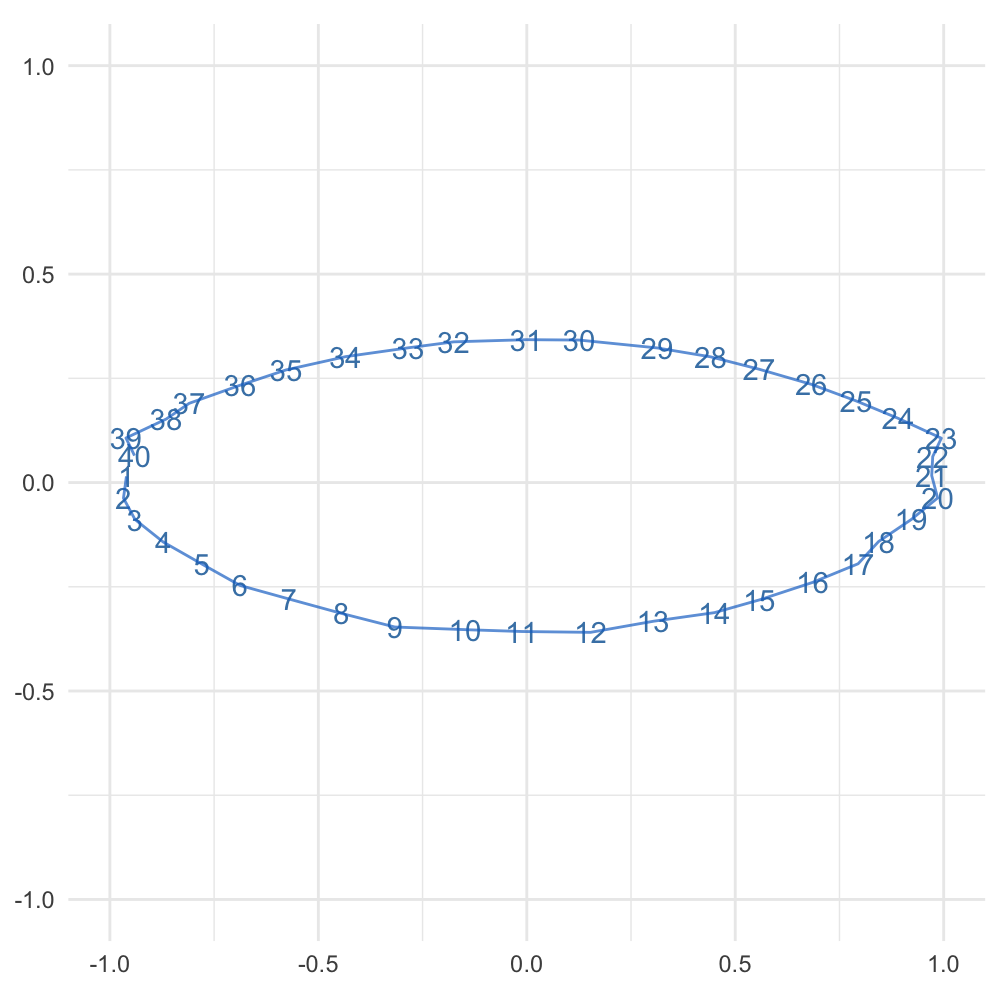}  \\
		\hline
	\end{tabular}
	\caption{Patterns obtained by MDS using squared distances associated with SW, Max-SW, and Max-D-SW with $p=2$ and $l=200$ projection directions.}
	\label{tabla2}
\end{table}

\noindent As in the case of $p=1$, as the sample size increases the figures are closer to the ``real patterns'' and in the case of ``small'' sample size, Max-D-SW gives better results.\\

\noindent In this example, we considered two-dimensional samples in order to give a more intuitive explanation of the behaviour of SW and its variants. Nonetheless, when increasing the ambient dimension while keeping the same type of distributional structure, similar results were obtained. Although the naive implementation of Max-D-SW also suffers from projection complexity, we encourage its use in medium-dimensional problems. Indeed, as discussed in \cite{nadjahi}, the quality of the empirical Wasserstein estimator deteriorates as the dimension increases, and even in dimension 10 the estimator may perform poorly for Gaussian distributions.

\subsection{Max-D-SW for pattern detection with edge-detected images of an object}
We also repeat the experiment with edge-detected images from the COIL-100 dataset, following the setup considered in \cite{sutherlandsamples}. The COIL-100 dataset contains color images of 100 objects, with 72 photographs of each object obtained by rotating it through 360 degrees and taking one photograph every 5 degrees.\\

\begin{figure}[htbp]
	\centering
	\begin{subfigure}[b]{0.45\textwidth} 
		\includegraphics[width=\textwidth]{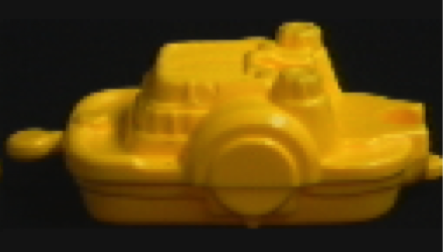}
		\caption{Original image of the submarine.}
		\label{fig:subfigA1}
	\end{subfigure}
	\hfill 
	\begin{subfigure}[b]{0.45\textwidth}
		\includegraphics[width=\textwidth]{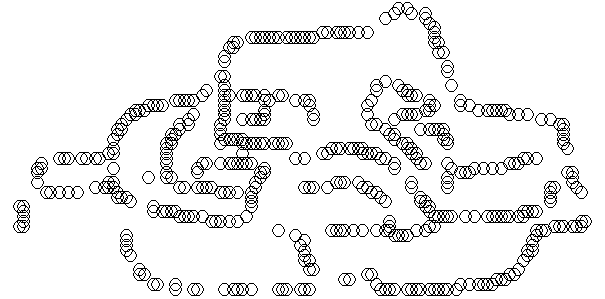}
		\caption{Edge-detected image of the submarine.}
		\label{fig:subfigB2}
	\end{subfigure}	
	\medskip 
	\begin{subfigure}[b]{0.45\textwidth}
		\includegraphics[width=\textwidth]{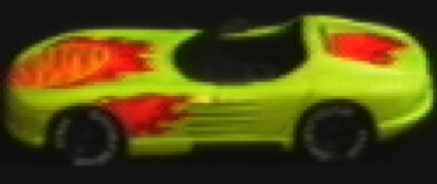}
		\caption{Original image of the car.}
		\label{fig:subfigC1}
	\end{subfigure}
	\hfill
	\begin{subfigure}[b]{0.45\textwidth}
		\includegraphics[width=\textwidth]{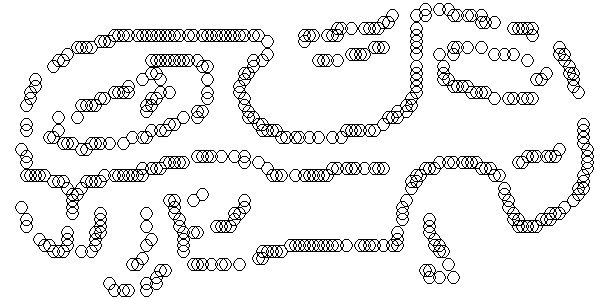}
		\caption{Edge-detected image of the car.}
		\label{fig:subfigD1}
	\end{subfigure}
	\caption{Original images of the submarine and car, together with their corresponding edge-detected images.}
	\label{objetos}
\end{figure}

\noindent We used the function \texttt{cannyEdges} from the \texttt{imager} R package to obtain $800$ data points from the detected edge of each object. As mentioned by the authors \cite{sutherlandsamples}, this problem is difficult when considering only the detected edges of the object and not the entire image. We added one more difficulty to the problem: instead of only identifying the rotation movement of a single object, we want the algorithm to find the rotation pattern in two objects and to classify the objects. Figure \ref{objetos} presents the original images of the two objects together with their corresponding edge-detected images. The edge-detected images show that the problem is difficult even for the human eye. 

\begin{figure}[htbp]
	\centering
	\begin{subfigure}[b]{0.32\textwidth} 
		\includegraphics[width=\textwidth]{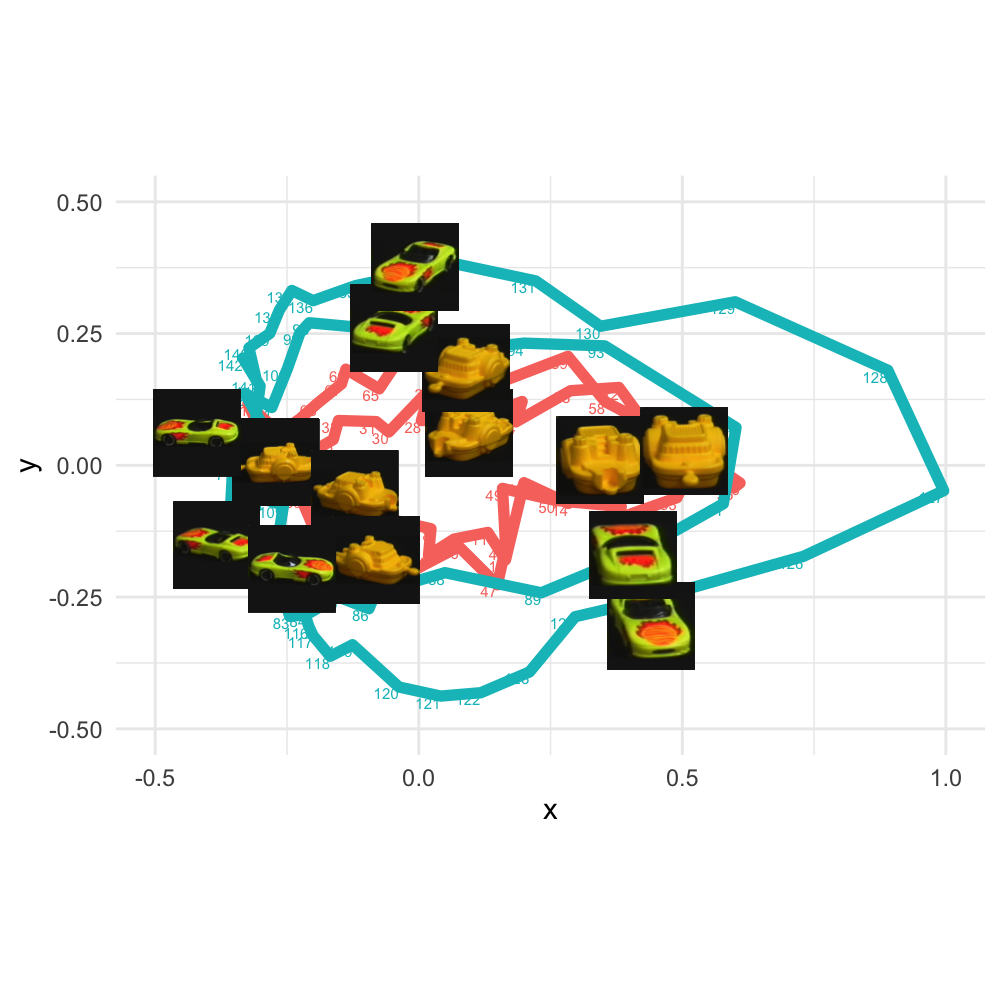}
		\caption{Sliced Wasserstein}
		\label{fig:subfigA}
	\end{subfigure}
	\hfill 
	\begin{subfigure}[b]{0.32\textwidth}
		\includegraphics[width=\textwidth]{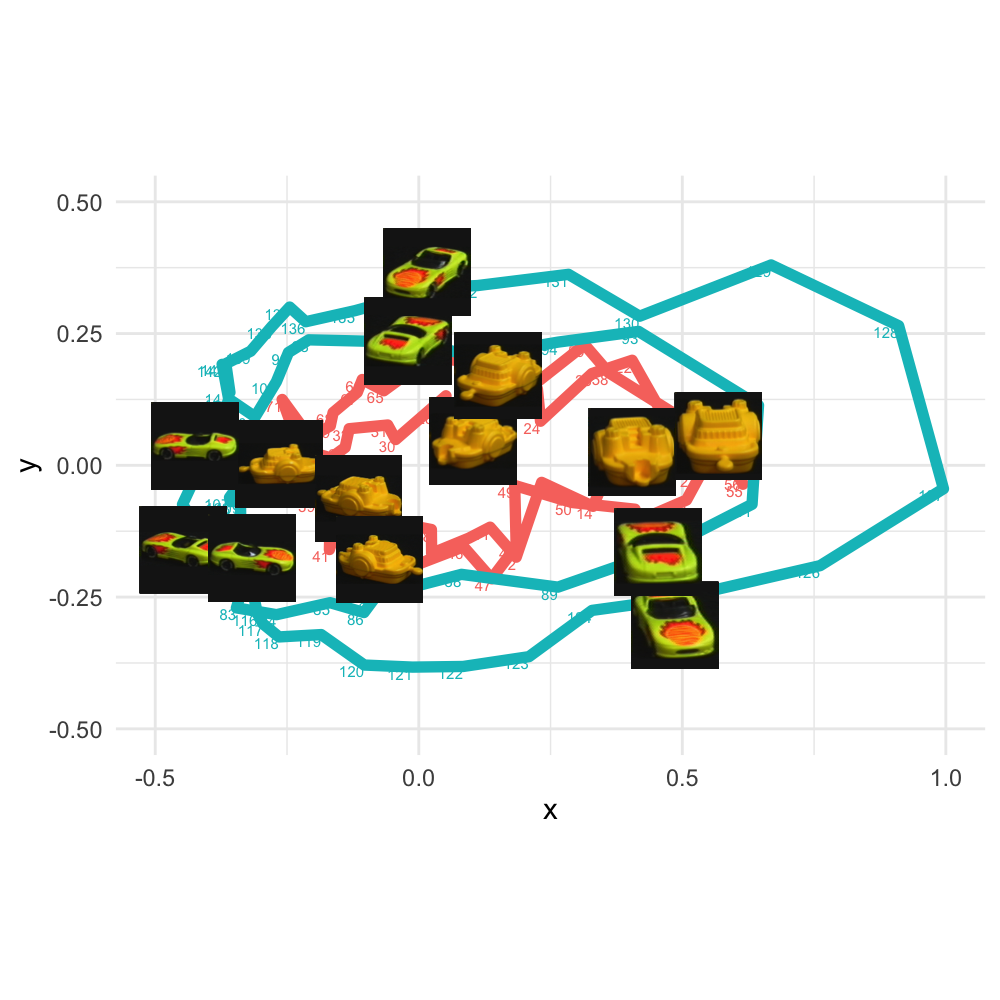}
		\caption{Max Sliced Wasserstein}
		\label{fig:subfigB}
	\end{subfigure}	
		\hfill 
	\begin{subfigure}[b]{0.32\textwidth}
		\includegraphics[width=\textwidth]{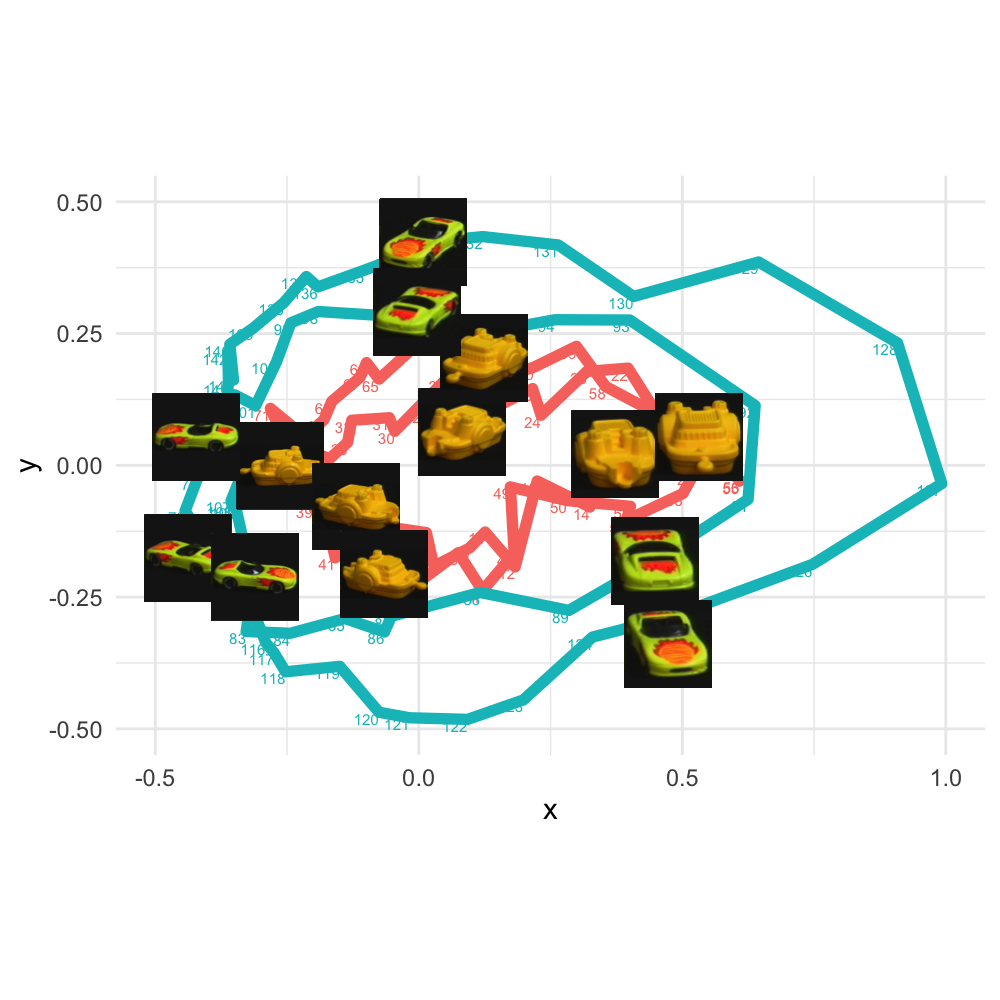}
		\caption{Max-D-Sliced Wasserstein}
		\label{fig:subfigC}
	\end{subfigure}	
	\medskip 
	\begin{subfigure}[b]{0.3\textwidth}
		\includegraphics[width=\textwidth]{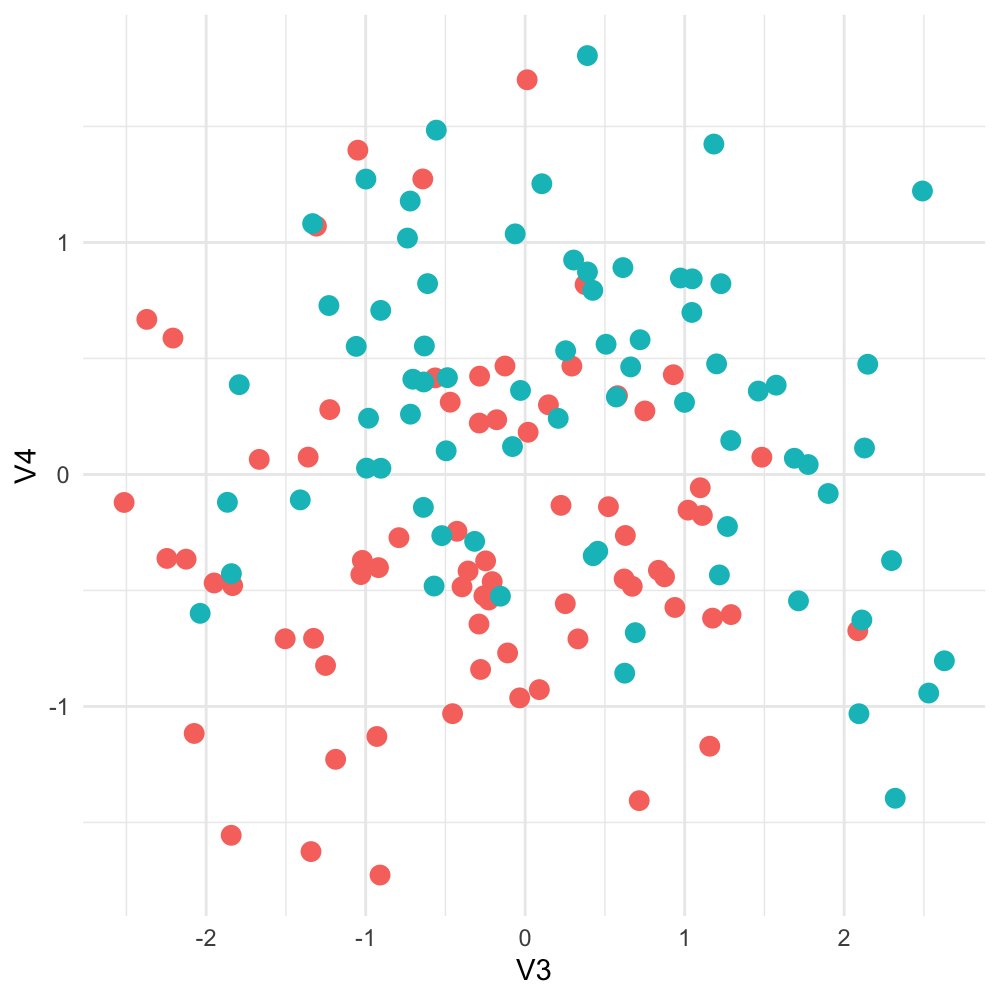}
		\caption{Sliced Wasserstein}
		\label{fig:subfigD}
	\end{subfigure}
	\hfill
	\begin{subfigure}[b]{0.3\textwidth}
		\includegraphics[width=\textwidth]{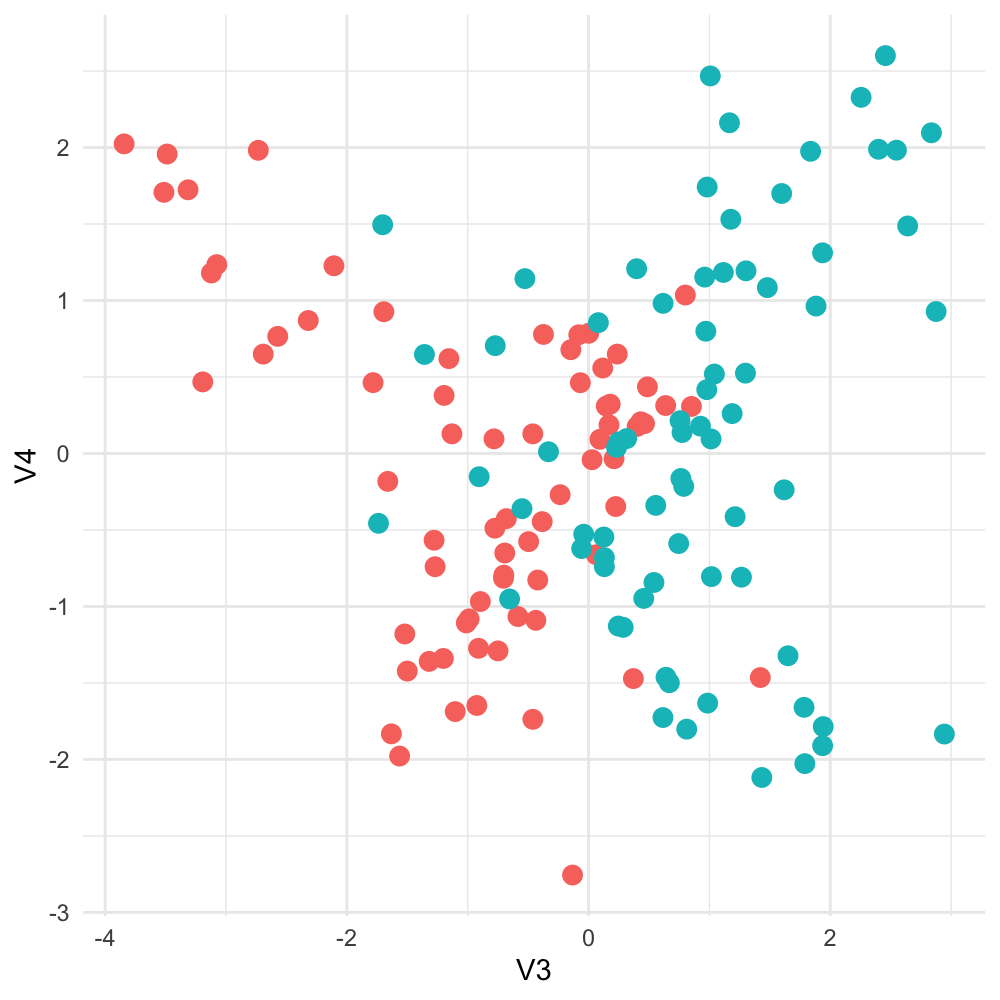}
		\caption{Max Sliced Wasserstein}
		\label{fig:subfigE}
	\end{subfigure}
		\hfill
	\begin{subfigure}[b]{0.33\textwidth}
		\includegraphics[width=\textwidth]{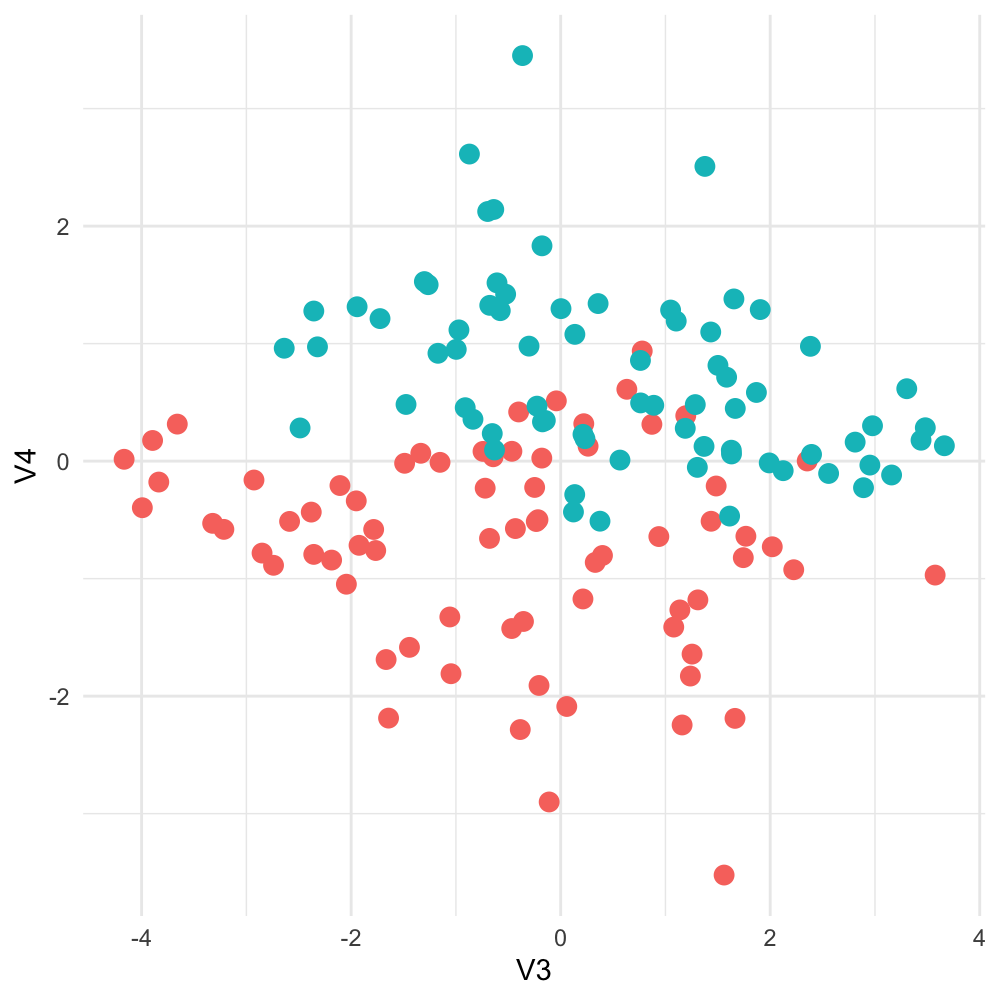}
		\caption{Max-D-Sliced Wasserstein}
		\label{fig:subfigF}
	\end{subfigure}
	\caption{First and second MDS coordinates (top row), and third and fourth MDS coordinates (bottom row), obtained using different distances.}
	\label{res_mds}
\end{figure}

\noindent The result obtained through multidimensional scaling is presented in Figure \ref{res_mds}. For all the distances we use $l=200$ projection directions. The top row shows the configurations obtained from the first two MDS coordinates for each distance. All methods capture the motion of both objects. Although they distinguish between the front and back orientations, resulting in two circles for each object rather than one, they provide a clear geometric representation of the rotational motion. In the bottom row, the third and fourth dimensions are presented. In this representation, the separation between the two objects is clearer when the Max-D-SW distance is used. 

\section{Conclusion}
\noindent Max-D-SW is the full-dimensional member of the Max-K-SW family and, in this sense, uses the largest number of orthogonal projection directions. At the same time, it preserves several useful features of sliced Wasserstein-type distances, such as favorable sample-complexity bounds, a simple naive implementation, and relatively low computational cost.  We showed that the most suitable distance for a problem depends on the desired task and the characteristics of the distribution. In particular, when the interest is to construct an  informative manifold embedding, when there is more than one important pattern or difference on the data or when the task is to find a pattern on a group of (almost) heavy-tailed samples, Max-D-SW seems to give better results even when using a naive implementation of the method.\\

\noindent \textbf{Acknowledgements}\\
Arturo Jaramillo Gil was supported by
the grant CBF2023-2024-2088.

\bibliographystyle{plain}
\bibliography{ref}

\end{document}